\documentclass{article}

\usepackage{arxiv}
\usepackage[round,authoryear]{natbib}

\usepackage[utf8]{inputenc}
\usepackage[T1]{fontenc}
\usepackage{hyperref}
\usepackage{url}
\usepackage{booktabs}
\usepackage{graphicx}
\usepackage{amsfonts}
\usepackage{amsmath}
\usepackage{amssymb}
\usepackage{nicefrac}
\usepackage{microtype}
\usepackage{xcolor}
\usepackage{tikz}
\usepackage{pgfplots}
\usepackage{multirow}
\usepackage{enumitem}
\usepackage{subcaption}
\usepackage{placeins}
\usepackage{fontawesome5}
\usepackage{amsthm}

\newtheorem{proposition}{Proposition}
\newtheorem{remark}{Remark}
\newtheorem{assumption}{Assumption}

\usetikzlibrary{arrows.meta, positioning, calc, fit, backgrounds, decorations.pathreplacing, shapes.geometric}
\pgfplotsset{compat=1.18}

\newcommand{\vth}{V_{\mathrm{th}}}

\newcommand{\reals}{\mathbb{R}}

\DeclareMathOperator{\softplus}{softplus}
\DeclareMathOperator{\sigmoid}{sigmoid}

\tikzset{
  snnbox/.style={draw, rounded corners=2pt, minimum width=2.4cm, minimum height=0.55cm,
    align=center, font=\scriptsize, fill=blue!8, thick},
  attnbox/.style={draw, rounded corners=2pt, minimum width=2.4cm, minimum height=0.55cm,
    align=center, font=\scriptsize, fill=orange!10, thick},
  iobox/.style={draw, rounded corners=2pt, minimum width=2.4cm, minimum height=0.55cm,
    align=center, font=\scriptsize, fill=green!8, thick},
  neuronbox/.style={draw, rounded corners=2pt, minimum width=1.6cm, minimum height=0.5cm,
    align=center, font=\scriptsize, fill=red!8, thick},
  projbox/.style={draw, rounded corners=1pt, minimum width=1.0cm, minimum height=0.4cm,
    align=center, font=\tiny, fill=gray!8},
  arr/.style={-{Stealth[length=1.8mm]}, thick, color=black!70},
  darr/.style={-{Stealth[length=1.8mm]}, thick, densely dashed, color=black!50},
  relarr/.style={-{Stealth[length=1.8mm,width=1.35mm]}, line width=0.55pt,
    opacity=0.62, shorten >=0.8pt, shorten <=0.8pt},
  lbl/.style={font=\tiny, text=black!60},
}
\title{NeuronSpark: Selective Spiking State Space Models\\with Adaptive Computation for Language Modeling}

\author{
  Zhengzheng Tang
  \And Hao Zhu
  \And Zhewei Chen
  \And Dajiang Tao
}

\renewcommand{\shorttitle}{NeuronSpark: Selective Spiking State Space Models}
\date{}
\hypersetup{hidelinks, pdftitle={NeuronSpark: Selective Spiking State Space Models with Adaptive Computation for Language Modeling}, pdfauthor={Zhengzheng Tang; Hao Zhu; Zhewei Chen; Dajiang Tao}}

\begin{document}

\maketitle
\raggedbottom

% ============================================================
% ABSTRACT
% ============================================================
\begin{abstract}
Spiking neural networks promise brain-inspired neural dynamics, yet native large-scale spiking language-model pretraining remains largely unexplored relative to conversion, distillation, and backbone-adaptation approaches.
We diagnose a central obstacle: standard spiking neurons apply the same firing sensitivity to every token, creating dense- or silent-firing regimes that are difficult to resolve with globally tuned parameters alone.
This problem is particularly acute in language modeling, where a single sequence contains tokens of vastly different informational roles, making a uniform firing regime poorly matched to the data.
We address this with \textsc{NeuronSpark}, a self-attention-free language model built around input-dependent spiking membrane dynamics, where each neuron's decay rate, gain, and firing threshold adapt to the current token content.
Together with gated associative memory and adaptive temporal aggregation, this selectivity enables billion-scale pretraining from random initialization without backbone transfer, conversion, or distillation.
We relate the neuron to fixed-parameter spiking models and the memory to linear attention.
At 1.1B parameters and 2.1B tokens, \textsc{NeuronSpark} reaches a 40.0 eight-task zero-shot average, exceeds comparable-budget Pythia/Mamba checkpoints, retains 95.2\%/93.2\% of their fully trained scores, and spontaneously develops a fast--slow hierarchy.
At 2,048 tokens, parameter-matched VHDL points reduce operation-energy proxy by 90.1\%/67.0\% and cycles by 94.1\%/80.3\% versus Transformer/Mamba; a physical FPGA path reaches 24.000\,s TTFT and 0.082440 TPS.
\end{abstract}

% ============================================================
% 1. INTRODUCTION
% ============================================================
\section{Introduction}

Language models built on Transformer attention or modern state space models achieve strong performance, but their computation is built on continuous, real-valued activations and dense matrix operations.
Spiking neural networks offer a fundamentally different substrate: computation through membrane-state dynamics, thresholded firing, and reset-driven recurrence rather than dense attention over token pairs.
We study this neural-dynamics mechanism together with its hardware realization: parameter-matched VHDL accounts operation-energy proxy and cycles, while a physical FPGA path measures TTFT and decode throughput under fixed protocols.
If a large language model could be built natively from such membrane dynamics, trained directly from scratch rather than converted or approximated from an existing model, it would open a path toward brain-inspired language processing grounded in neural state evolution rather than Transformer imitation.

That goal has proven stubbornly difficult.
Prior spiking language models approach this goal only partially: SpikeGPT~\citep{zhu2023spikegpt} explores spiking generative modeling at smaller scale with RWKV-inspired recurrence; SpikeBERT~\citep{lv2023spikebert} and SpikeLM~\citep{xing2024spikelm} distil from BERT; SpikingMamba~\citep{huang2026spikingmamba} grafts spikes onto a pretrained Mamba model; and the SpikingSSMs model~\citep{shen2025spikessms} uses a pretrained surrogate dynamic network to bypass the neuron's iterative training loop.
These workarounds suggest a shared structural obstacle that has not been isolated at the neuron-dynamics level.
The core problem is that a standard spiking neuron applies the same firing sensitivity to every input token, regardless of content.
In language modeling, a noun, a punctuation mark, and a syntactic function word carry vastly different information, yet the same fixed threshold governs how all of them are processed.
At the start of training, when the model has no established representations, this mismatch caused severe instability in our runs: neurons either fire at nearly every token and collapse representations into a constant signal, or fire almost never and stop receiving any useful gradient.
Language models face this heterogeneity at every step of every sequence, making a single globally adjusted firing threshold an especially brittle control mechanism.
In our development runs, this manifests as divergent or stalled training unless the neuron itself is made selective.

We call this \emph{firing-regime rigidity} and treat it as a testable diagnosis: making the neuron's firing sensitivity \emph{input-dependent} directly targets the observed dense- and silent-firing failure modes.
When each token can modulate the neuron's own decay rate, gain, and threshold, different tokens naturally stabilize different firing regimes within the same layer at the same training step, stabilizing the observed training dynamics from the start.
Input-dependent neuron dynamics additionally close an expressiveness gap at inference, since fixed membrane parameters lack neuron-level content-dependent transition control that modern selective sequence models depend on~\citep{gu2023mamba}, which helps explain why many successful spiking LMs borrow selectivity from a non-spiking or pretrained backbone rather than placing it in the neuron itself.
Addressing both problems with one change makes SelectivePLIF a compact extension of prior spiking neurons for native language-model pretraining.
Long-range integration uses two further components: a gated associative memory for content-addressable retrieval without softmax attention, and adaptive temporal aggregation over short internal processing steps per token.
Together, these form a self-attention-free architecture grounded in neural membrane dynamics.

We introduce \textsc{NeuronSpark}, instantiating this design at 1.1B parameters and pretraining it from random initialization on 2.1B tokens, making it, to our knowledge, the first billion-parameter spiking language model trained without backbone transfer, conversion, or distillation.
We present it as a scientific feasibility study: the primary claim is that firing-regime rigidity has a tractable solution and that the resulting architecture is trainable and evaluable at meaningful scale.
Our contributions are:
\begin{itemize}[nosep, leftmargin=1.5em]
  \item \textbf{Firing-regime rigidity}: a diagnosis of why fixed-parameter spiking dynamics can collapse into dense or silent regimes during language pretraining, and why token-level input-dependence directly addresses this failure (§\ref{sec:prelim}, §\ref{sec:architecture_stability}).
  \item \textbf{SelectivePLIF and its formal properties}: an input-dependent spiking neuron that subsumes standard fixed-parameter neurons as special cases and empirically addresses both training stability and content-dependent filtering (Proposition~\ref{prop:hierarchy}).
  \item \textbf{A grounded associative memory}: a PLIF-gated memory module formally connected to the linear-attention literature, providing content-addressable accumulation without softmax self-attention (Remark~\ref{rem:linattn}).
  \item \textbf{\textsc{NeuronSpark}-1.1B and hardware realization}: to our knowledge, the first natively pretrained billion-parameter spiking LM; its 40.0 average, emergent fast--slow dynamics, parameter-matched VHDL costs, and FPGA timing establish trainability, structure, and realizable execution within stated boundaries.
\end{itemize}

% ---- FIGURE 1: COMBINED ARCHITECTURE ----
\begin{figure}[t]
\centering
\includegraphics[width=\linewidth]{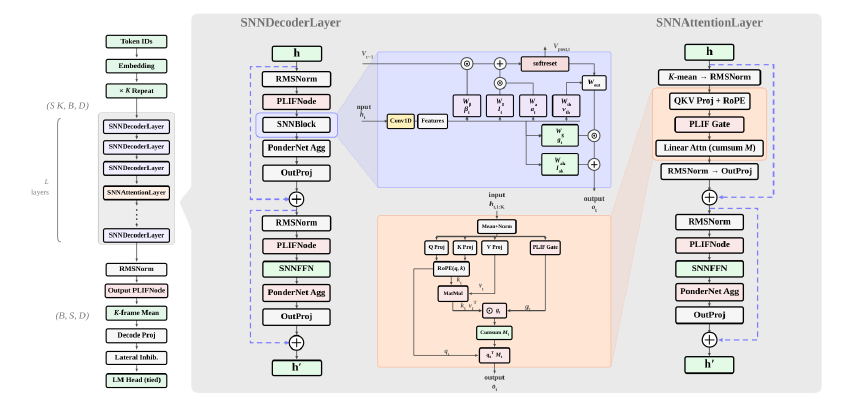}
\caption{\textbf{Architecture of \textsc{NeuronSpark}.}
Left: token embeddings are repeated over $K$ frames, processed by mixed SNN layers, and decoded through an output PLIF neuron, $K$-frame aggregation, and a tied LM head.
Right: the expanded layer view shows the SNNDecoderLayer and SNNAttentionLayer, the selective SNNBlock membrane recurrence, and the PLIF-gated associative-memory path used in place of conventional self-attention.}
\label{fig:architecture}
\end{figure}

% ============================================================
% 2. RELATED WORK
% ============================================================
\section{Related work}
\noindent\textbf{Selective sequence models.}
S4~\citep{gu2022efficiently} introduced structured state space models for long-range sequence modeling; Mamba~\citep{gu2023mamba} added input-dependent (\emph{selective}) state transitions by making the continuous SSM transition matrix a function of the current input, achieving Transformer-competitive language modeling and establishing that LTI recurrences lack input-dependent transitions for content filtering.
Follow-on work includes Mamba-2~\citep{dao2024transformers}, RetNet~\citep{sun2023retentive}, RWKV~\citep{peng2023rwkv}, GLA~\citep{yang2024gated}, and the fast-weight linear-attention family~\citep{katharopoulos2020transformers,schlag2021linear} from which our associative memory is derived (Remark~\ref{rem:linattn}).
All of these operate on continuous real-valued states and none addresses how to introduce the equivalent of input-dependent selectivity into discrete spiking membrane dynamics.
\textsc{NeuronSpark} bridges this gap: where Mamba conditions the state-transition matrix on the input, SelectivePLIF conditions the membrane threshold, decay rate, and gain on the current token, achieving selectivity through neuron-level dynamics rather than a continuous state equation.

\noindent\textbf{Spiking language models.}
SNNs trained via surrogate gradients~\citep{neftci2019surrogate,fang2021incorporating} have been applied to language through several strategies: spiking recurrent or backbone-adapted generative models (SpikeGPT~\citep{zhu2023spikegpt}, SpikingMamba~\citep{huang2026spikingmamba}, Sorbet~\citep{tang2025sorbet}) and distillation or conversion from pretrained ANNs (SpikeBERT~\citep{lv2023spikebert}, SpikeLM~\citep{xing2024spikelm}, BrainGPT~\citep{tang2024braingpt}, SpikeLLM~\citep{xing2024spikellm}, SpikingLLM~\citep{guo2025spikingllmcssa,xu2025spikingllmconversion}, \citealt{shi2025howtogetspikingllms,yuan2025easyspiking}).
Across these strategies, the spiking neurons generally remain fixed-parameter or inherit selectivity from the surrounding architecture (RWKV recurrence, Mamba transitions, Transformer attention), rather than making the membrane parameters themselves token-dependent.
\textsc{NeuronSpark} inverts this relationship: by making the neuron itself selective, native billion-scale pretraining from random initialization becomes tractable in our setting without relying on a pretrained backbone as the source of selectivity.

\noindent\textbf{Spiking state-space sequence models.}
SPikE-SSM~\citep{zhong2024spikessm} and P-SpikeSSM~\citep{bal2025pspikessm} demonstrate that spiking dynamics can support long-range \emph{sequence classification}; SpikingSSMs~\citep{shen2025spikessms} extends this to word-level language modeling on WikiText-103 and is the closest prior benchmark to ours.
Critically, the SpikingSSMs model requires a separately pre-trained Surrogate Dynamic Network (SDN) to bypass the iterative LIF recurrence during training: the SDN is trained on synthetic spike data before the main model, then frozen and used to predict spike trains in parallel, sidestepping the neuron's training dynamics entirely.
This pre-training stage is necessary precisely because the LIF neurons in SpikingSSMs are fixed-parameter; the globally learnable threshold it adopts controls the average firing rate but does not provide the per-token firing selectivity needed for language modeling.
\textsc{NeuronSpark} requires no such bypass: SelectivePLIF's input-dependent parameters resolve the instability at the neuron level, enabling end-to-end training from random initialization without any auxiliary pre-training stage, surrogate network, or separate fitting procedure.

% ============================================================
% 3. METHOD
% ============================================================
\begingroup
\section{Method}
Standard spiking language models apply the same membrane decay and firing threshold to every token, making the isolated neuron an LTI system that lacks content-dependent filtering~\citep{gu2023mamba} and, in our development runs, tends toward two degenerate regimes: dense firing, where representations collapse toward a constant signal, and silence, where surrogate gradients vanish. A globally learnable threshold controls only the average firing rate and is brittle across heterogeneous token distributions. Both failures point to input-independent neuron parameters; the following sections show how making decay, gain, and threshold functions of the current token closes the expressiveness gap and stabilizes our reported training run.

\subsection{Preliminaries}
\label{sec:prelim}

\noindent\textbf{Standard PLIF baseline.}
The PLIF neuron~\citep{fang2021incorporating} maintains membrane potential $V_t \in \reals$ via
\begin{equation}
\label{eq:plif_base}
  V_t = \beta \cdot V_{t-1} + (1-\beta) \cdot I_t, \qquad
  s_t = \Theta(V_t - \vth), \qquad
  V_t \leftarrow V_t - \vth \cdot s_t,
\end{equation}
where $I_t \in \reals$ is the input current, $\beta = \sigmoid(w) \in (0,1)$ is a learned but \emph{input-independent} temporal decay, $\vth > 0$ is a fixed threshold, $s_t \in \{0,1\}$ is the binary spike, and the third expression implements soft reset.
In the backward pass, $\Theta$ is replaced by the sigmoid surrogate gradient~\citep{neftci2019surrogate}.
Because $\beta$ and $\vth$ are constant, this dynamics is a linear time-invariant (LTI) recurrence, identical to an LTI SSM with decay $\beta$ and gain $(1-\beta)$.

\noindent\textbf{LTI expressiveness gap.}
\citet{gu2023mamba} showed that LTI sequence models lack input-dependent transitions for content filtering: all input positions receive the same state-transition regardless of content.
Standard PLIF inherits this limitation.
Spiking LMs built on fixed-parameter PLIF neurons~\citep{lv2023spikebert, zhu2023spikegpt, zhong2024spikessm} therefore lack neuron-level content-selective transitions, even when surrounding architectures add other forms of selectivity.

\noindent\textbf{Firing-regime rigidity: diagnostic statement.}
The introduction characterized the attractor problem informally; here we state the diagnostic quantity used throughout the paper.
Define the \emph{layer firing rate} as $\rho_{\ell,s} = \frac{1}{K}\sum_k s_{s,k}^{(\ell)} \in [0,1]$.
With fixed $\vth$, the expected firing rate $\mathbb{E}[\rho_{\ell,s}]$ depends only on the magnitude distribution of the pre-spike membrane, determined by weight initialization and current embedding scale, neither of which is controlled during training.
Both $\rho_{\ell,s} \to 1$ (saturation: residual stream collapses toward a constant-fire signal) and $\rho_{\ell,s} \to 0$ (silence: surrogate gradient $\partial s/\partial V$ vanishes) are empirically stable failure modes in our fixed-parameter development runs, because the optimizer receives only a global threshold-control signal.
With input-dependent $\vth(t)$, the neuron can maintain distinct firing rates for different tokens in the same forward pass, avoiding these failure modes in our runs.
This is supported in §\ref{sec:architecture_stability}: ablated variants that remove selective modulation or its stabilizing controls fail to reach stable convergence under the tested recipe, while SelectivePLIF converges stably.

\noindent\textbf{Our Target.}
Let $\mathbf{h}_t \in \reals^D$ denote the token representation at time step $t$.
The goal of Sections~\ref{sec:selective_plif}--\ref{sec:memory_layer} is to instantiate a neuron model of the form
\begin{equation}
\label{eq:open_target}
  V_t = \beta(\mathbf{h}_t) \cdot V_{t-1} + \alpha(\mathbf{h}_t) \cdot I(\mathbf{h}_t), \qquad
  s_t = \Theta\!\big(V_t - \vth(\mathbf{h}_t)\big), \qquad
  V_t \leftarrow V_t - \vth(\mathbf{h}_t) \cdot s_t,
\end{equation}
where $\beta(\cdot)$, $\alpha(\cdot)$, $I(\cdot)$, and $\vth(\cdot)$ are all learnable input-dependent maps, such that:
\begin{enumerate}[nosep, leftmargin=1.8em, label=(\roman*)]
  \item \eqref{eq:open_target} contains \eqref{eq:plif_base} and scalar LTI SSMs as special cases (Proposition~\ref{prop:hierarchy}),
  \item a PLIF-gated associative memory provides content-addressable accumulation grounded in the linear-attention recurrence (Remark~\ref{rem:linattn}),
  \item the resulting architecture converges stably at billion-parameter scale under the tested recipe and assumptions below.
\end{enumerate}

\begin{assumption}[Soft reset preserves suprathreshold information]
\label{asm:softreset}
Subtracting $\vth \cdot s_t$ rather than resetting $V_t$ to zero preserves the suprathreshold excess $V_t - \vth$ in subsequent membrane integration.
We assume this excess is informative and that discarding it (hard reset) degrades downstream task performance.
This assumption is supported by §\ref{sec:architecture_stability}: the \emph{continuous health control} ablation removes the mechanism we use to avoid hard-reset discontinuities and fails to reach stable convergence under the tested recipe.
\end{assumption}

\begin{assumption}[$K$ frames are sufficient for temporal depth]
\label{asm:kframes}
We assume that $K{=}12$ temporal frames per token are sufficient to capture the temporal depth required by the evaluated language-modeling tasks.
This is supported by §\ref{sec:neural_temporal_dynamics}: the empirical distribution of soft frame centers spans $[3.0, 6.7]$ expected frame index, well within the $K{=}12$ budget, with no saturation at the boundary.
\end{assumption}

\begin{assumption}[PLIF-gated cumulative memory is sufficient for evaluated tasks]
\label{asm:memory}
We assume that the scalar-modulated outer-product memory $M_T = \sum_t g_t \mathbf{k}_t \mathbf{v}_t^\top$ (§\ref{sec:memory_layer}) provides sufficient associative recall capacity for the eight zero-shot evaluation tasks.
This is supported by §\ref{sec:architecture_stability}: removing the associative-memory pathway causes training instability under the tested recipe, and no alternative long-range pathway was found to be trainable in our development budget.
\end{assumption}

\subsection{Model components}
\label{sec:model_components}

All three components below are instantiations of the same primitive, the SelectivePLIF dynamics of \eqref{eq:open_target}, in different operational configurations.
\textsc{NeuronSpark} follows a three-stage pipeline: token embeddings are repeated $K$ times along the temporal axis to form a $(S{\cdot}K, B, D)$ tensor, processed by a stack of $L$ mixed SNN layers, then aggregated back to $(S, B, D)$ and projected to vocabulary logits (Figure~\ref{fig:architecture}).

\noindent\textbf{Selective PLIF neuron.}
\label{sec:selective_plif}
Standard PLIF~\citep{fang2021incorporating} maintains membrane potential $V_t$ with input-independent decay $\beta$ and threshold $\vth$, which constitutes an LTI recurrence that \citet{gu2023mamba} showed is insufficient for content-based reasoning.
We make all parameters input-dependent.
Given projected input $\mathbf{h}_t \in \reals^D$, SelectivePLIF computes:
\begingroup
\small
\begin{equation}
\label{eq:sel_dyn}
\begin{aligned}
\beta_t &= \sigmoid\!\big(\mathbf{W}_\beta \mathbf{h}_t + \mathbf{b}_\beta\big),\quad
\alpha_t = \softplus\!\big(\mathbf{W}_\alpha \mathbf{h}_t + \mathbf{b}_\alpha\big),\quad
\vth(t) = V_{\min} + \big|\mathbf{W}_{\mathrm{th}} \mathbf{h}_t + \mathbf{b}_{\mathrm{th}}\big|,\\
V_t &= \beta_t \cdot V_{t-1} + \alpha_t \cdot I_t,\quad
s_t = \Theta\!\big(V_t - \vth(t)\big),\quad
V_t \leftarrow V_t - \vth(t)\cdot s_t,
\end{aligned}
\end{equation}
\endgroup
where $I_t,\beta_t,\alpha_t,\vth(t),V_t\in\reals^{DN}$ after expanding each model channel into $N$ hidden banks, and the third expression is a soft reset preserving suprathreshold information (Assumption~\ref{asm:softreset}).
High $\beta_t$ preserves memory; low $\beta_t$ enables rapid adaptation; dynamic $\vth(t)$ controls firing selectivity, with all three quantities gated by token content.

\begin{proposition}[Special-case hierarchy]
\label{prop:hierarchy}
\begin{enumerate}[nosep, leftmargin=1.8em, label=(\roman*)]
  \item Setting $\mathbf{W}_\beta{=}\mathbf{W}_\alpha{=}\mathbf{W}_{\mathrm{th}}{=}0$ with appropriate biases recovers standard PLIF~\eqref{eq:plif_base} exactly.
  \item Further taking $\vth(t)\to\infty$ (so $s_t\equiv 0$) reduces the membrane update to $V_t = \beta V_{t-1} + (1{-}\beta)I_t$, recovering a scalar LTI SSM.
\end{enumerate}
Thus LTI-style scalar recurrence and standard PLIF are recovered as special cases of SelectivePLIF.
% Proof in Appendix~\ref{app:prop_proof}.
\end{proposition}

Fixed-parameter spiking LMs instantiate a subset of SelectivePLIF's parameter space in the sense of Proposition~\ref{prop:hierarchy}; \textsc{NeuronSpark} uses the full selective parameterization at billion-parameter scale.

\noindent\textbf{SNN decoder layer.}
\label{sec:decoder_layer}
Each decoder layer deploys SelectivePLIF along the token-sequence temporal axis.
The selective recurrent block (SNNBlock) reads post-convolution features $\mathbf{x}_t$, computes the four modulation projections of \eqref{eq:sel_dyn}, and exposes a gated continuous readout:
\begingroup
\small
\begin{equation}
\label{eq:snnblock}
\begin{aligned}
V^{\mathrm{hid}}_{\mathrm{post},t}
&= \mathrm{SelectivePLIF}\!\big(\mathbf{W}_{\mathrm{in}}\mathbf{x}_t,\;\beta_t,\;\alpha_t,\;\vth(t)\big),\\[-0.1em]
\mathbf{o}_t
&= \mathbf{W}_{\mathrm{out}} V^{\mathrm{hid}}_{\mathrm{post},t}
   \odot \sigmoid(\mathbf{W}_g\mathbf{x}_t) + \mathbf{W}_{\mathrm{sk}}\mathbf{x}_t.
\end{aligned}
\end{equation}
\endgroup
A second sublayer uses two fixed-parameter PLIF branches (the $\beta_t\!\equiv\!\beta$ special case of Proposition~\ref{prop:hierarchy}(i)) as gate and up activations for a spiking FFN, keeping the residual stream continuous.
The $K$ framewise outputs are aggregated by differentiable PonderNet-style soft temporal weights~\citep{banino2021pondernet}:
\begin{equation}
\label{eq:ponder_main}
\lambda_{s,k} = p_{s,k}\!\prod_{j<k}(1-p_{s,j}),\quad p_{s,k} = \sigmoid(\mathbf{w}_{\mathrm{halt}}^\top\mathbf{z}_{s,k}),\quad
\bar{\mathbf{z}}_s = \textstyle\frac{\sum_{k=1}^{K}\lambda_{s,k}\,\mathbf{z}_{s,k}}{\sum_{k=1}^{K}\lambda_{s,k}+\varepsilon},
\end{equation}
where $\bar{\mathbf{z}}_s$ is broadcast back across all $K$ frames before residual addition.
This accumulation is the empirical counterpart to integration depth $\tau{=}{-1}/\!\log\beta_t$; Assumption~\ref{asm:kframes} and \S\ref{sec:neural_temporal_dynamics} support the adequacy of $K{=}12$ for the evaluated tasks.
Full residual equations are in Appendix~\ref{app:arch_details}.

\noindent\textbf{Associative memory layer.}
\label{sec:memory_layer}
Every $m$-th layer ($m{=}4$) deploys SelectivePLIF in its gate-axis configuration: a fixed-parameter PLIF neuron computes a scalar write coefficient $g_t$ controlling how strongly information enters the cumulative associative store.
After averaging the $K$ frames per token, the layer applies RMSNorm, QKV projections with RoPE~\citep{su2024roformer}, and the gated write:
\begin{equation}
\label{eq:memory}
g_t = \tfrac{1}{D}\mathbf{1}^\top\!\big((1{-}\beta_g)\odot V_{\mathrm{post},t}^{(g)}\big),\qquad
M_T = \textstyle\sum_{t=1}^{T} g_t\,\mathbf{k}_t\mathbf{v}_t^\top,\qquad
\mathbf{o}_T = \mathbf{q}_T^\top M_T,
\end{equation}
where $\mathbf{q}_T,\mathbf{k}_T \in \reals^{D_k}$ are $L_2$-normalized, $\mathbf{v}_T \in \reals^{D_v}$, and $M_T \in \reals^{D_k\times D_v}$.

\begin{remark}[Linear-attention grounding]
\label{rem:linattn}
When $g_t\equiv 1$, \eqref{eq:memory} reduces exactly to the standard cumulative outer-product memory of linear attention~\citep{katharopoulos2020transformers,schlag2021linear}.
The PLIF-derived $g_t$ provides input-dependent signed modulation within the fast-weight family, rather than an architecturally novel module introduced by analogy alone.
\end{remark}

The scalar coefficient $g_t$ is not a probability; it gives signed write modulation analogous to hippocampal gated encoding~\citep{teyler1986hippocampal,hasselmo1994laminar,nakazawa2002ca3}, and supports Assumption~\ref{asm:memory}: removing this pathway causes training instability under the tested recipe (\S\ref{sec:architecture_stability}).
Output-boundary details are in Appendix~\ref{app:arch_details}; initialization and gradient-compensation details are in Appendix~\ref{app:train_details}.
\endgroup

% ============================================================
% 4. EXPERIMENTS
% ============================================================
\section{Experiments}

\subsection{Experimental setup}
\label{sec:setup}

We report the current 24-layer, 1.16B-parameter model as \textsc{NeuronSpark}-1.1B.
The model uses $D{=}1024$, $D_{\mathrm{ff}}{=}3072$, maximum temporal frames $K{=}12$, state expansion $N{=}8$, a 64{,}002-token vocabulary, and PLIF-gated associative-memory layers every fourth layer.
The reported checkpoint is trained on 2.1B tokens with next-token prediction and is evaluated with lm-evaluation-harness under the same zero-shot protocols as the public Transformer and SSM baselines.
The 2.1B-token budget reflects an expired rented compute allocation rather than training saturation; hardware evaluation uses a vocabulary-aligned derivative of the same 24-layer backbone and is separately scoped in Appendix~\ref{app:exp_setup}.
For ARC-E/C, PIQA, and OBQA we report \texttt{acc\_norm}; for WinoGrande, BoolQ, MMLU, and C-Eval we report \texttt{acc}.
Full model and optimization details are deferred to Appendix~\ref{app:exp_setup}.

The experiments follow a single evidence chain rather than independent score reporting.
We first test whether the architecture reaches a recognizable same-scale language-model capability band, then ask whether this capability is obtained more effectively than comparable-token Transformer and Mamba checkpoints, whether the training recipe is stable enough to scale, whether the recorded internal dynamics actually use token- and layer-dependent temporal computation, and finally whether a bit-exact realization reduces operation-energy proxy and inference time under explicit VHDL and FPGA boundaries.
Raw benchmark values, rejected-run definitions, hardware measurements, and neural-dynamics statistics are reported in Appendix~\ref{app:details}.

\subsection{Benchmark results}
\label{sec:main_results}

\begin{table}[!htbp]
  \centering
  \begin{minipage}{\linewidth}
    \vspace{0pt}
    \centering
    \scriptsize
    \renewcommand{\arraystretch}{1.35}
    \setlength{\tabcolsep}{2.3pt}
    \textbf{Main same-scale language-model baselines}\\[0.15em]
    \resizebox{\linewidth}{!}{%
    \begin{tabular}{llccccccccc}
      \toprule
      Model & Tokens & ARC-E & ARC-C & Wino. & BoolQ & PIQA & OBQA & MMLU & C-Eval & Avg \\
      \midrule
      Pythia-1B (step1000) & 2.1B & 0.305 & 0.212 & 0.488 & 0.392 & 0.546 & 0.236 & 0.233 & 0.241 & 0.332 \\
      Pythia-1B & 300B & 0.489 & 0.265 & 0.532 & 0.607 & 0.694 & 0.312 & 0.232 & 0.233 & 0.420 \\
      Mamba-1.6B (pre-1000) & 3.67B & 0.335 & 0.214 & 0.493 & 0.398 & 0.548 & 0.250 & 0.230 & 0.226 & 0.337 \\
      Mamba-1.6B & 300B & 0.487 & 0.270 & 0.545 & 0.624 & 0.714 & 0.318 & 0.240 & 0.233 & 0.429 \\
      \textsc{NeuronSpark}-1.1B & 2.1B & 0.462 & 0.247 & 0.511 & 0.601 & 0.604 & 0.284 & 0.232 & 0.262 & 0.400 \\
      OPT-1.3B & 300B & 0.509 & 0.278 & 0.590 & 0.571 & 0.726 & 0.328 & 0.249 & 0.230 & 0.435 \\
      TinyLlama-1.1B & 3000B & 0.559 & 0.311 & 0.582 & 0.582 & 0.729 & 0.356 & 0.256 & 0.246 & 0.453 \\
      Mamba-790M & 300B & 0.539 & 0.288 & 0.571 & 0.610 & 0.724 & 0.344 & 0.234 & 0.237 & 0.443 \\
      \bottomrule
    \end{tabular}
    }
  \end{minipage}
  \vspace{0.45em}

  \begin{minipage}{\linewidth}
    \vspace{0pt}
    \centering
    \scriptsize
    \renewcommand{\arraystretch}{1.25}
    \setlength{\tabcolsep}{1.7pt}
    \textbf{SNN language-model references}\\[0.15em]
    \resizebox{\linewidth}{!}{%
    \begin{tabular}{lllcccccccccc}
      \toprule
      Model & Size & Tokens & WT2$\downarrow$ & ARC-E & ARC-C & Wino. & BoolQ & PIQA & Hella. & OBQA & HeadQA & SST-2 \\
      \midrule
      SpikeGPT & 216M & 5.0B & 280.51 & 0.316 & 0.230 & 0.498 & 0.387 & 0.548 & 0.258 & 0.258 & 0.254 & 0.522 \\
      SpikingLLM-CSSA$^\dagger$, $T{=}2$ & 1.3B & 10.0B & -- & 0.403 & 0.215 & 0.530 & 0.529 & 0.631 & 0.410 & 0.241 & 0.238 & -- \\
      SpikingLLM-CSSA$^\dagger$, $T{=}4$ & 1.3B & 10.0B & -- & 0.408 & 0.222 & 0.543 & 0.534 & 0.642 & 0.426 & 0.247 & 0.241 & -- \\
      WTA-SpikingFormer$^\ddagger$ & 0.4B & -- & -- & -- & -- & 0.502 & -- & 0.534 & 0.259 & -- & -- & 0.890 \\
      \textsc{NeuronSpark} & 1.1B & 2.1B & 9.70 & 0.462 & 0.247 & 0.511 & 0.601 & 0.604 & 0.284 & 0.284 & 0.223 & 0.493 \\
      \bottomrule
    \end{tabular}
    }
    \vspace{0.12em}
    \begin{minipage}{0.98\linewidth}
    \scriptsize\raggedright
    $^\dagger$CSSA/SKD protocol details are provided in Appendix~\ref{app:eval_columns}.\\[-0.05em]
    $^\ddagger$WD reports causal-LM commonsense; WE reports encoder-only SST-2.
    \end{minipage}
  \end{minipage}
  \vspace{0.15em}
  \caption{\textbf{Zero-shot benchmark results.}}
  \label{tab:benchmark_results}
\end{table}

\textsc{NeuronSpark}-1.1B demonstrates that neural dynamics can serve as the core substrate of a billion-parameter language-model backbone, and scores above the selected public Transformer and Mamba checkpoints trained under comparable early-token budgets.
Table~\ref{tab:benchmark_results} shows a 40.0 eight-task average after 2.1B tokens, above Pythia-1B at the matched step1000 checkpoint (33.2) and Mamba-1.6B at the closest released pre-1000 checkpoint despite its larger 3.67B-token budget (33.7).
Against fully trained 300B-token baselines, the same score reaches 95.2\% of Pythia-1B and 93.2\% of Mamba-1.6B, while retaining task-level anchors such as near parity to Pythia-1B on BoolQ (0.601 versus 0.607), the same MMLU score as Pythia-1B (0.232), and the strongest C-Eval score among listed baselines (0.262 versus at most 0.246).
These results should be read as zero-shot capability evidence at the reported 2.1B-token training budget; hardware realization is evaluated separately in Section~\ref{sec:hardware_realization} under its own parameter and measurement controls.
Against SNN language-model references, the main comparison is on shared downstream evaluations: \textsc{NeuronSpark}-1.1B is higher than the CSSA/SKD SpikingLLM rows~\citep{guo2025spikingllmcssa} on ARC-E, ARC-C, BoolQ, and OBQA, while WT2 perplexity is listed as an additional language-modeling diagnostic for the official SpikeGPT checkpoint~\citep{zhu2023spikegpt}.
WTA-SpikingFormer~\citep{zhou2026wtaspikingformer} is retained as a heterogeneous SNN reference because its row combines causal WD commonsense scores with encoder-only WE SST-2.
This purely neural-dynamical backbone is feasible and competitive against the listed generative SNN language-model references on common downstream evaluations; the next section reports token-budget comparisons as public-checkpoint evidence rather than a fully data-controlled architecture attribution.

\subsection{Token efficiency}
\label{sec:token_efficiency}

\begin{figure}[!htbp]
  \centering
  \begin{minipage}[t]{0.49\linewidth}
    \vspace{0pt}
    \centering
    \includegraphics[width=\linewidth]{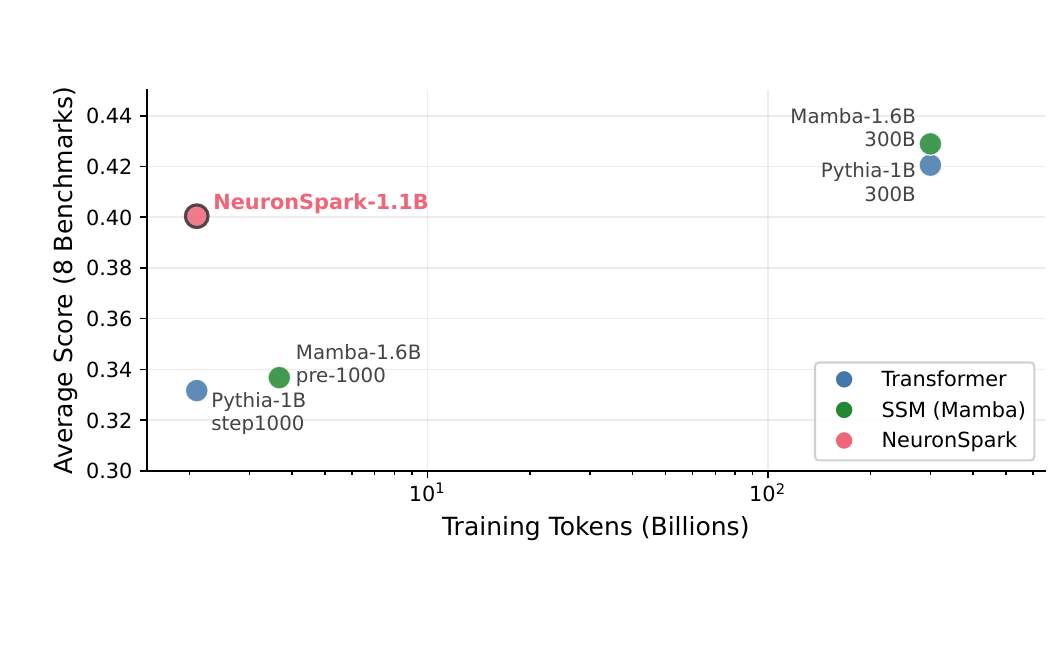}
  \end{minipage}\hfill
  \begin{minipage}[t]{0.49\linewidth}
    \vspace{0pt}
    \centering
    \includegraphics[width=\linewidth]{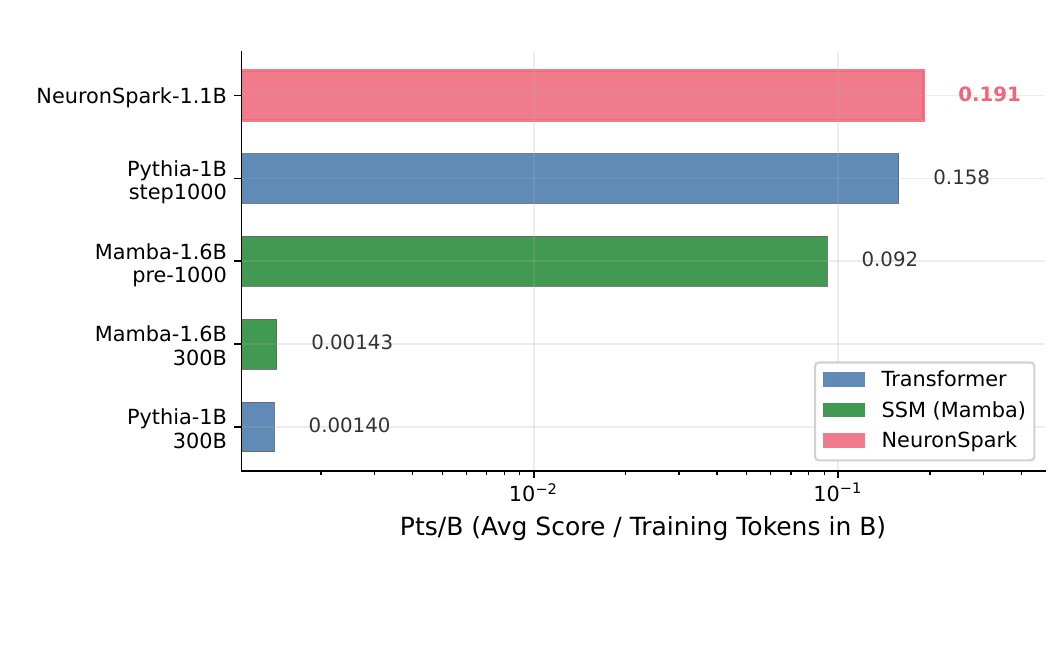}
  \end{minipage}
  \caption{\textbf{Token-efficiency views.}}
  \label{fig:token_efficiency_views}
\end{figure}

Under comparable released early-token checkpoints, \textsc{NeuronSpark}-1.1B has higher measured Pts/B than the selected Transformer and Mamba references, with 0.191 versus 0.158 for Pythia-1B (step1000) and 0.092 for Mamba-1.6B (pre-1000).
The corresponding raw scores are 40.0 after 2.1B tokens, compared with 33.2 for Pythia-1B at 2.1B tokens and 33.7 for Mamba-1.6B at 3.67B tokens.
Fully trained 300B-token baselines provide mature-model capability references, where \textsc{NeuronSpark}-1.1B reaches 95.2\% of Pythia-1B and 93.2\% of Mamba-1.6B.
We use these full-budget rows as mature-model capability anchors, while token efficiency is evaluated on comparable released early checkpoints.
Figure~\ref{fig:token_efficiency_views} visualizes these same-budget and full-budget comparisons by separating score--budget position from the resulting Pts/B ranking.
These comparisons evaluate the final model as a whole, while architectural attribution requires testing whether its mechanisms can be independently perturbed without breaking training.
The architecture-stability ablations perturb modulation compensation, input-dependent PLIF parameterization, health-control, and cross-token pathways before full pretraining.

\subsection{Architecture stability}
\label{sec:architecture_stability}

\begin{figure}[!htbp]
  \centering
  \includegraphics[width=\linewidth]{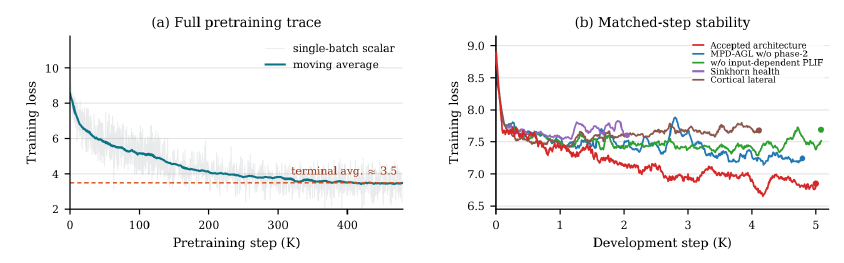}
  \caption{\textbf{Training-loss and matched-step stability diagnostics.}}
  \label{fig:training_diagnostics}
\end{figure}

\begin{figure}[t]
  \centering
  \includegraphics[width=0.99\linewidth,trim=0 20pt 0 0,clip]{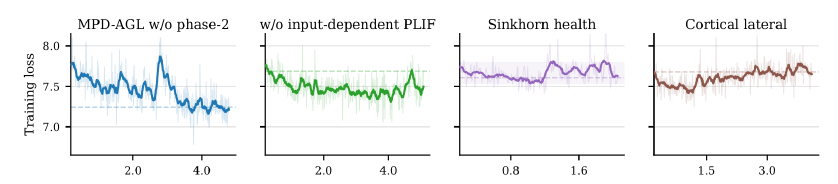}
  {\scriptsize
  \setlength{\tabcolsep}{2.0pt}
  \renewcommand{\arraystretch}{0.90}
  \begin{tabular*}{0.99\linewidth}{@{\extracolsep{\fill}}lccc}
    \toprule
    Variant & End step & Best $\rightarrow$ final loss & Rejection signal \\
    \midrule
    MPD-AGL w/o phase-2 & 4.79K & 6.78 @ 3.76K $\rightarrow$ 7.24 & Reversal \\
    w/o input-dependent PLIF & 5.08K & 7.06 @ 3.83K $\rightarrow$ 7.69 & Reversal \\
    Sinkhorn health & 2.05K & 7.41 @ 1.19K $\rightarrow$ 7.60 & High-loss band \\
    Cortical lateral & 4.11K & 7.25 @ 1.19K $\rightarrow$ 7.68 & Reversal \\
    \bottomrule
  \end{tabular*}
  }
  \caption{\textbf{Post-warmup architecture-stability ablation traces and matched-step summary.}}
  \label{fig:ablation_detail_grid}
\end{figure}

Under our tested recipe, stable training was observed only when mechanisms operated as a coupled dynamical system, and removing tested components prevented convergence in development runs.
The architecture is therefore a stability-driven combination strategy rather than a blind accumulation of mechanisms.
Figure~\ref{fig:training_diagnostics}(a,b) shows the contrast: the accepted architecture continues to a moving-average loss near 3.5, whereas rejected variants either burst, rise after partial descent, oscillate in limit cycles, or fail under lateral propagation.
Figure~\ref{fig:ablation_detail_grid} assigns these failures to four tested components, covering phase-2 modulation compensation, input-dependent PLIF parameterization, continuous health control, and the associative-memory pathway, and summarizes the matched-step rejection signals below the traces.
Convergence ablations do not expose every mechanism, including the single-token $K$-frame multi-step accumulation.
For this mechanism, the relevant check is whether a forward pass exhibits membrane-state evolution rather than repeated feed-forward computation.

\subsection{Neural temporal dynamics}
\label{sec:neural_temporal_dynamics}

The extra $K$-frame computation is membrane-state integration rather than a uniform repeated feed-forward loop, as the recorded output PLIF trace contains 51 spike events over 168 internal frames in the displayed channel.
The recorded trace in Figure~\ref{fig:neural_dynamics_v2}(a) keeps continuous post-reset membrane voltage and leakage-derived outputs, matching the actual signal type passed through the model rather than a pure binary spike stream.
Since the $K$-frame forward trace shows continuous membrane-state evolution rather than repeated computation, we further test whether these states are used uniformly or acquire content- and layer-dependent temporal weights.

Within the fixed integer budget of $K{=}12$ executed frames, \textsc{NeuronSpark} learns a content-dependent temporal allocation, with content-bearing starts and subword fragments using later soft frame centers and high-frequency closed-class and surface tokens concentrating earlier.
The non-integer values are soft aggregation centers over the 12 executed frames, not fractional executed steps.
Across 64 TinyStories validation prompts, content word starts and subword pieces center later than articles/determiners, prepositions/conjunctions/particles, and punctuation/quotes (5.80 and 5.73 versus 3.84, 4.41, and 4.71).
Selective retention spans $\beta(t){=}0.42$ to 0.91 across tokens and SNNBlocks, with the per-token temporal-depth view deferred to Appendix~\ref{app:dynamics_details}.
This content-dependent temporal allocation provides token-level evidence that the learned dynamics organize computation over time, which raises a stronger test of whether pretraining separates slow state retention from fast memory modulation across depth and associative-memory pathways.

\noindent
\begin{minipage}{0.995\linewidth}
  \centering
  \includegraphics[width=\linewidth]{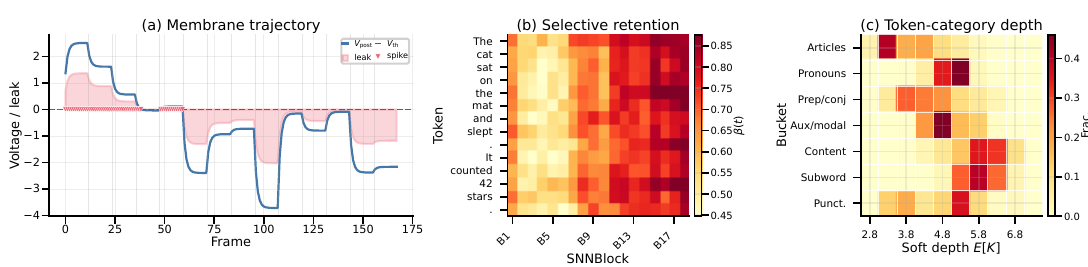}
	  \captionsetup{hypcap=false}
\captionof{figure}{\textbf{Neural temporal diagnostics.}}
	  \label{fig:neural_dynamics_v2}
    \label{fig:membrane_trajectory}
    \label{fig:ek_category_distribution}
\end{minipage}
\vspace{0.20em}

\begin{figure}[!htbp]
  \centering
  \includegraphics[width=0.92\linewidth]{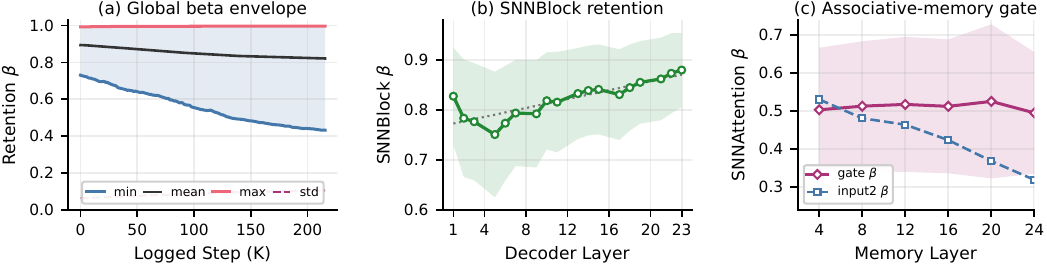}
  \captionsetup{hypcap=false}
  \caption{\textbf{Emergent fast--slow temporal organization in the logged pretraining trace.}}
  \label{fig:timescale_split}
\end{figure}

Pretraining spontaneously induces fast--slow timescale separation, with SNNBlocks forming progressively slower membrane retention across decoder depth while associative-memory gates remain in a faster modulation regime.
Figure~\ref{fig:timescale_split} shows the separation at the level needed for this claim: layer-mean SNNBlock timescales increase from roughly 3.5 to 7.8 recurrent frames, and final SNNBlock mean $\beta$ correlates with decoder depth (Pearson $r{=}0.86$).
The deepest SNN layers reach means around 0.87--0.88, while SNNAttention associative-memory gates remain centered near $\beta{\approx}0.50$ with broader dispersion, consistent with fast gated memory writes rather than slow residual retention.
Together with the benchmark, token-efficiency, and token-depth results, this shows that the multi-frame neural-dynamics design is not redundant computation, but a mechanism that supports efficient pretraining, content-dependent temporal allocation, and emergent fast--slow timescale separation.
The same dynamics expose fixed-threshold release and add-only accumulation paths, making hardware realization the final test of whether the learned mechanism yields lower execution cost.

\subsection{Hardware realization}
\label{sec:hardware_realization}

The optimized VHDL datapath reduces both simulated operation-energy and cycle costs against parameter-matched Transformer and Mamba implementations.

\noindent
\begin{minipage}[t]{0.485\linewidth}
  \vspace{0pt}
  \centering
  \includegraphics[width=\linewidth]{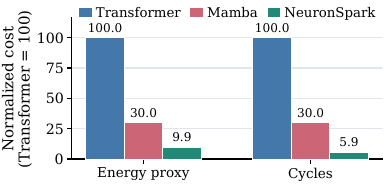}
  \captionsetup{hypcap=false}
  \captionof{figure}{\textbf{Same-parameter VHDL execution costs.}}
  \label{fig:vhdl_costs}
\end{minipage}\hfill
\begin{minipage}[t]{0.485\linewidth}
  \vspace{0pt}
  \centering
  \includegraphics[width=\linewidth]{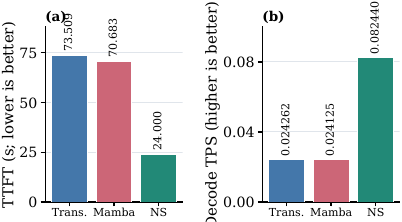}
  \captionsetup{hypcap=false}
  \captionof{figure}{\textbf{Measured FPGA inference speed.}}
  \label{fig:fpga_speed}
\end{minipage}

At 2,048 tokens, Figure~\ref{fig:vhdl_costs} reports 90.1\%/67.0\% lower operation-energy proxy at the energy point and 94.1\%/80.3\% fewer cycles at the speed point versus Transformer/Mamba; all datapaths pass bit-exact checks.
Figure~\ref{fig:fpga_speed} next tests physical-board timing.

The one-direction FPGA path reduces measured TTFT and increases decode throughput under the fixed short-context protocol.
On the ALINX VD100 (AMD VE2302), Figure~\ref{fig:fpga_speed} reports 24.000\,s TTFT and 0.082440 TPS: $3.063\times/2.945\times$ TTFT speedups and $3.398\times/3.417\times$ TPS gains over Transformer/Mamba, with 0.337\% maximum A/B spread.
The evidence chain is scoped to one-direction BF16, a two-token prompt, and four generated tokens; VHDL energy is a simulation proxy, not measured board joules.

% ============================================================
% 5. CONCLUSION
% ============================================================
\FloatBarrier
\section{Conclusion}

We introduced \textsc{NeuronSpark}, a self-attention-free language model whose token-dependent membrane decay, gain, and threshold realize selective state-space updates with periodic PLIF-gated memory.

At 1.1B parameters trained on 2.1B tokens, it scores 40.0 above selected comparable-budget Transformer/Mamba checkpoints and learns fast--slow allocation; VHDL lowers operation-energy/cycle proxies, and FPGA reaches 24.000\,s TTFT and 0.082440 TPS.
This establishes a trainable billion-scale substrate and realizable low-cost path, bounded by single-seed, non-data-controlled pretraining, proxy-only energy, and one-direction short-context FPGA measurements.

% ============================================================
% REFERENCES
% ============================================================
\bibliographystyle{plainnat}
\bibliography{references}

@article{gu2023mamba,
  title={Mamba: Linear-Time Sequence Modeling with Selective State Spaces},
  author={Gu, Albert and Dao, Tri},
  journal={arXiv preprint arXiv:2312.00752},
  year={2023}
}

@inproceedings{gu2022efficiently,
  title={Efficiently Modeling Long Sequences with Structured State Spaces},
  author={Gu, Albert and Goel, Karan and R{\'e}, Christopher},
  booktitle={International Conference on Learning Representations},
  year={2022}
}

@article{dao2024transformers,
  title={Transformers are {SSM}s: Generalized Models and Efficient Algorithms Through Structured State Space Duality},
  author={Dao, Tri and Gu, Albert},
  journal={arXiv preprint arXiv:2405.21060},
  year={2024}
}

@article{peng2023rwkv,
  title={{RWKV}: Reinventing {RNN}s for the Transformer Era},
  author={Peng, Bo and Alcaide, Eric and Anthony, Quentin and others},
  journal={arXiv preprint arXiv:2305.13048},
  year={2023}
}

@article{sun2023retentive,
  title={Retentive Network: A Successor to Transformer for Large Language Models},
  author={Sun, Yutao and Dong, Li and Huang, Shaohan and others},
  journal={arXiv preprint arXiv:2307.08621},
  year={2023}
}

@article{yang2024gated,
  title={Gated Linear Attention Transformers with Hardware-Efficient Training},
  author={Yang, Songlin and Wang, Bailin and Shen, Yikang and Panda, Rameswar and Kim, Yoon},
  journal={arXiv preprint arXiv:2312.06635},
  year={2024}
}

@inproceedings{katharopoulos2020transformers,
  title={Transformers are {RNN}s: Fast Autoregressive Transformers with Linear Attention},
  author={Katharopoulos, Angelos and Vyas, Apoorv and Pappas, Nikolaos and Fleuret, Fran{\c{c}}ois},
  booktitle={International Conference on Machine Learning},
  year={2020}
}

@article{schlag2021linear,
  title={Linear Transformers are Secretly Fast Weight Programmers},
  author={Schlag, Irwan and Irie, Kazuki and Schmidhuber, J{\"u}rgen},
  journal={arXiv preprint arXiv:2102.11174},
  year={2021}
}

@article{touvron2023llama,
  title={{LLaMA}: Open and Efficient Foundation Language Models},
  author={Touvron, Hugo and Lavril, Thibaut and Izacard, Gautier and others},
  journal={arXiv preprint arXiv:2302.13971},
  year={2023}
}

@inproceedings{biderman2023pythia,
  title={{Pythia}: A Suite for Analyzing Large Language Models Across Training and Scaling},
  author={Biderman, Stella and Schoelkopf, Hailey and Anthony, Quentin Gregory and others},
  booktitle={International Conference on Machine Learning},
  year={2023}
}

@inproceedings{zhang2019root,
  title={Root Mean Square Layer Normalization},
  author={Zhang, Biao and Sennrich, Rico},
  booktitle={Advances in Neural Information Processing Systems},
  year={2019}
}

@article{su2024roformer,
  title={{RoFormer}: Enhanced Transformer with Rotary Position Embedding},
  author={Su, Jianlin and Ahmed, Murtadha and Lu, Yu and Pan, Shengfeng and Bo, Wen and Liu, Yunfeng},
  journal={Neurocomputing},
  volume={568},
  pages={127063},
  year={2024}
}

@article{zhu2023spikegpt,
  title={{SpikeGPT}: Generative Pre-trained Language Model with Spiking Neural Networks},
  author={Zhu, Rui-Jie and Zhao, Qihang and Li, Guoqi and Eshraghian, Jason K},
  journal={arXiv preprint arXiv:2302.13939},
  year={2023}
}

@article{lv2023spikebert,
  title={{SpikeBERT}: A Language Spikformer Learned from {BERT} with Knowledge Distillation},
  author={Lv, Changze and Li, Tianlong and Xu, Jianhan and Gu, Chenxi and Ling, Zixuan and Zhang, Cenyuan and Zheng, Xiaoqing and Huang, Xuanjing},
  journal={arXiv preprint arXiv:2308.15122},
  year={2023}
}

@inproceedings{xing2024spikelm,
  title={{SpikeLM}: Towards General Spike-Driven Language Modeling via Elastic Bi-Spiking Mechanisms},
  author={Xing, Xingrun and Zhang, Zheng and Ni, Ziyi and Xiao, Shitao and Ju, Yiming and Fan, Siqi and Wang, Yequan and Zhang, Jiajun and Li, Guoqi},
  booktitle={Proceedings of the 41st International Conference on Machine Learning},
  series={Proceedings of Machine Learning Research},
  volume={235},
  pages={54698--54714},
  publisher={PMLR},
  year={2024}
}

@inproceedings{xing2024spikellm,
  title={{SpikeLLM}: Scaling up Spiking Neural Network to Large Language Models via Saliency-based Spiking},
  author={Xing, Xingrun and Gao, Boyan and Liu, Zheng and Clifton, David A. and Xiao, Shitao and Zhang, Wanpeng and Du, Li and Zhang, Zheng and Li, Guoqi and Zhang, Jiajun},
  booktitle={International Conference on Learning Representations},
  year={2025}
}

@article{tang2024braingpt,
  title={{BrainGPT}: A Brain-Inspired {SNN}-Based Large Language Model},
  author={Tang, Zhengzheng and Zhu, Eva},
  journal={OpenReview},
  year={2024}
}

@inproceedings{tang2025sorbet,
  title={Sorbet: A Neuromorphic Hardware-Compatible Transformer-Based Spiking Language Model},
  author={Tang, Kaiwen and Yan, Zhanglu and Wong, Weng-Fai},
  booktitle={International Conference on Machine Learning},
  year={2025}
}

@article{zhong2024spikessm,
  title={{SPikE-SSM}: A Sparse, Precise, and Efficient Spiking State Space Model for Long Sequences Learning},
  author={Zhong, Yan and Zhao, Ruoyu and Wang, Chao and Guo, Qinghai and Zhang, Jianguo and Lu, Zhichao and Leng, Luziwei},
  journal={arXiv preprint arXiv:2410.17268},
  year={2024}
}

@inproceedings{shen2025spikessms,
  title={{SpikingSSMs}: Learning Long Sequences with Sparse and Parallel Spiking State Space Models},
  author={Shen, Shuaijie and Wang, Chao and Huang, Renzhuo and Zhong, Yan and Guo, Qinghai and Lu, Zhichao and Zhang, Jianguo and Leng, Luziwei},
  booktitle={Proceedings of the AAAI Conference on Artificial Intelligence},
  volume={39},
  number={19},
  pages={20380--20388},
  year={2025},
  doi={10.1609/aaai.v39i19.34245}
}

@inproceedings{bal2025pspikessm,
  title={{P-SpikeSSM}: Harnessing Probabilistic Spiking State Space Models for Long-Range Dependency Tasks},
  author={Bal, Malyaban and Sengupta, Abhronil},
  booktitle={International Conference on Learning Representations},
  year={2025}
}

@article{guo2025spikingllmcssa,
  title={{SpikingLLM}: Spiking Large Language Models with Causal Spiking Self-Attention and Spike-Form Knowledge Distillation},
  author={Guo, Sihang and Zhou, Chenlin and Wang, Jiaqi and Chen, Kehai and Meng, Qingyan and Ma, Zhengyu},
  journal={OpenReview},
  year={2025}
}

@article{xu2025spikingllmconversion,
  title={{SpikingLLM}: A Conversion-Based Method with Window Inhibition Mechanism for Spiking Large Language Models},
  author={Xu, Zekai and Li, Hanqi and You, Kang and Nie, Chen and Guo, Qinghai and Wang, Xiang and He, Zhezhi},
  journal={OpenReview},
  year={2025}
}

@article{shi2025howtogetspikingllms,
  title={How to Get Spiking {LLM}s? A Dual {ANN}-to-{SNN} Conversion with Layer-Wise Calibration},
  author={Shi, Wanli and Fang, Haoran and Fang, Jinjie and Zhang, Haozhen and You, Runwen and Chen, Tianshuo and Liu, Zhaogeng and Chang, Yi and Gu, Bin},
  journal={OpenReview},
  year={2025}
}

@article{yuan2025easyspiking,
  title={EasySpiking: Spike-Friendly Function Approximations for Spiking {LLM}s Without Fine-Tuning},
  author={Yuan, Xinzhe and Peng, Xiang and Xiong, Huan and Gu, Bin},
  journal={OpenReview},
  year={2025}
}

@article{huang2026spikingmamba,
  title={{SpikingMamba}: Towards Energy-Efficient Large Language Models via Knowledge Distillation from Mamba},
  author={Huang, Yulong and Tang, Jianxiong and Wang, Chao and Wang, Ziyi and Zhang, Jianguo and Lu, Zhichao and Cheng, Bojun and Leng, Luziwei},
  journal={Transactions on Machine Learning Research},
  year={2026},
  note={Accepted by TMLR}
}

@article{zhou2026wtaspikingformer,
  title={Winner-Take-All Spiking Transformer for Language Modeling},
  author={Zhou, Chenlin and Guo, Sihang and Wang, Jiaqi and Ma, Dongyang and Che, Kaiwei and Chen, Baiyu and Meng, Qingyan and Ma, Zhengyu and Tian, Yonghong},
  journal={arXiv preprint arXiv:2604.11321},
  year={2026}
}

@article{neftci2019surrogate,
  title={Surrogate Gradient Learning in Spiking Neural Networks: Bringing the Power of Gradient-based Optimization to Spiking Neural Networks},
  author={Neftci, Emre O and Mostafa, Hesham and Zenke, Friedemann},
  journal={IEEE Signal Processing Magazine},
  volume={36},
  number={6},
  pages={51--63},
  year={2019}
}

@article{fang2021incorporating,
  title={Incorporating Learnable Membrane Time Constants to Enhance Learning of Spiking Neural Networks},
  author={Fang, Wei and Yu, Zhaofei and Chen, Yanqi and Masquelier, Timoth{\'e}e and Huang, Tiejun and Tian, Yonghong},
  journal={arXiv preprint arXiv:2007.05785},
  year={2021}
}

@inproceedings{fang2024parallel,
  title={Parallel Spiking Neurons with High Efficiency and Ability to Learn Long-term Dependencies},
  author={Fang, Wei and Yu, Zhaofei and Zhou, Zhaokun and Chen, Ding and Chen, Yanqi and Ma, Zhengyu and Masquelier, Timoth{\'e}e and Tian, Yonghong},
  booktitle={Advances in Neural Information Processing Systems},
  year={2023}
}

@article{banino2021pondernet,
  title={{PonderNet}: Learning to Ponder},
  author={Banino, Andrea and Balaguer, Jan and Blundell, Charles},
  journal={arXiv preprint arXiv:2107.05407},
  year={2021}
}

@article{teyler1986hippocampal,
  title={The Hippocampal Memory Indexing Theory},
  author={Teyler, Timothy J and DiScenna, Pascal},
  journal={Behavioral Neuroscience},
  volume={100},
  number={2},
  pages={147--154},
  year={1986}
}

@article{hasselmo1994laminar,
  title={Laminar Selectivity of the Cholinergic Suppression of Synaptic Transmission in Rat Hippocampal Region {CA1}: Computational Modeling and Brain Slice Physiology},
  author={Hasselmo, Michael E and Schnell, Eric},
  journal={Journal of Neuroscience},
  volume={14},
  number={6},
  pages={3898--3914},
  year={1994}
}

@article{nakazawa2002ca3,
  title={Requirement for Hippocampal {CA3} {NMDA} Receptors in Associative Memory Recall},
  author={Nakazawa, Kazu and Quirk, Michael C and Chitwood, Raymond A and Watanabe, Masahiko and Yeckel, Mark F and Sun, Linus D and Kato, Akira and Carr, Candice A and Johnston, Daniel and Wilson, Matthew A and Tonegawa, Susumu},
  journal={Science},
  volume={297},
  number={5579},
  pages={211--218},
  year={2002}
}

% ============================================================
% APPENDIX
% ============================================================
\clearpage
\appendix

\section{Architectural and training details}
\label{app:arch_details}

\subsection{Full decoder-layer residual equations}

For token index $s \in \{1, \dots, S\}$ and frame index $k \in \{1, \dots, K\}$, let $\mathbf{h}_{s,k}$ denote the incoming residual state.
The first decoder sublayer applies boundary conversion, framewise selective spiking recurrence, token-level aggregation, output projection, and feature-centering before broadcasting the same token-level residual back across all $K$ frames:
\begin{equation}
\label{eq:app_decoder_block}
\begin{aligned}
\mathbf{a}^{(1)}_{s,k} &= \phi_{\mathrm{PLIF}}\!\big(\mathrm{RMSNorm}(\mathbf{h}_{s,k})\big), \\
\mathbf{z}^{\mathrm{blk}}_{s,k} &= \mathrm{SNNBlock}\!\big(\mathbf{a}^{(1)}_{s,k}\big), \\
\bar{\mathbf{z}}^{\mathrm{blk}}_{s} &= \frac{\sum_{k=1}^{K} \lambda^{\mathrm{blk}}_{s,k} \, \mathbf{z}^{\mathrm{blk}}_{s,k}}{\sum_{k=1}^{K}\lambda^{\mathrm{blk}}_{s,k}+\varepsilon}, \\
\hat{\mathbf{z}}^{\mathrm{blk}}_{s} &= \mathbf{W}^{\mathrm{blk}}_{\mathrm{out}} \bar{\mathbf{z}}^{\mathrm{blk}}_{s}
  - \frac{1}{D}\mathbf{1}\mathbf{1}^{\top} \mathbf{W}^{\mathrm{blk}}_{\mathrm{out}} \bar{\mathbf{z}}^{\mathrm{blk}}_{s}, \\
\mathbf{h}^{(1)}_{s,k} &= \mathbf{h}_{s,k} + \hat{\mathbf{z}}^{\mathrm{blk}}_{s}.
\end{aligned}
\end{equation}
where $\phi_{\mathrm{PLIF}}(\cdot)$ denotes the fixed-PLIF boundary conversion defined in \eqref{eq:app_fixed_plif}, $\mathbf{z}^{\mathrm{blk}}_{s,k}$ is the framewise output of the selective recurrent block, $\lambda^{\mathrm{blk}}_{s,k}$ is the training-time temporal aggregation weight from \eqref{eq:app_ponder}, $\bar{\mathbf{z}}^{\mathrm{blk}}_{s} \in \reals^{D}$ is the token-level aggregated representation, and the centering term subtracts the feature mean before the residual is broadcast back to every frame of token $s$.

The second decoder sublayer applies the same residual pattern to the spiking FFN:
\begin{equation}
\label{eq:app_decoder_ffn}
\begin{aligned}
\mathbf{a}^{(2)}_{s,k} &= \phi_{\mathrm{PLIF}}\!\big(\mathrm{RMSNorm}(\mathbf{h}^{(1)}_{s,k})\big), \\
\mathbf{z}^{\mathrm{ffn}}_{s,k} &= \mathrm{SNNFFN}\!\big(\mathbf{a}^{(2)}_{s,k}\big), \\
\bar{\mathbf{z}}^{\mathrm{ffn}}_{s} &= \frac{\sum_{k=1}^{K} \lambda^{\mathrm{ffn}}_{s,k} \, \mathbf{z}^{\mathrm{ffn}}_{s,k}}{\sum_{k=1}^{K}\lambda^{\mathrm{ffn}}_{s,k}+\varepsilon}, \\
\hat{\mathbf{z}}^{\mathrm{ffn}}_{s} &= \mathbf{W}^{\mathrm{ffn}}_{\mathrm{out}} \bar{\mathbf{z}}^{\mathrm{ffn}}_{s}
  - \frac{1}{D}\mathbf{1}\mathbf{1}^{\top} \mathbf{W}^{\mathrm{ffn}}_{\mathrm{out}} \bar{\mathbf{z}}^{\mathrm{ffn}}_{s}, \\
\mathbf{h}'_{s,k} &= \mathbf{h}^{(1)}_{s,k} + \hat{\mathbf{z}}^{\mathrm{ffn}}_{s}.
\end{aligned}
\end{equation}
where $\mathbf{z}^{\mathrm{ffn}}_{s,k}$ is the framewise spiking-FFN output, $\lambda^{\mathrm{ffn}}_{s,k}$ is its temporal aggregation weight, and $\mathbf{h}'_{s,k}$ is the layer output passed to the next layer.

\subsection{Boundary conversion, SNNBlock, and spiking FFN}

Whenever a fixed-parameter PLIF neuron feeds a downstream module, the module consumes a leak activation rather than the raw membrane potential:
\begin{equation}
\label{eq:app_fixed_plif}
\begin{aligned}
V_t &= \beta \odot V_{t-1} + (1-\beta) \odot I_t, \\
s_t &= \Theta(V_t - \vth), \qquad
V_{\mathrm{post},t} = V_t - \vth \odot s_t, \\
\mathbf{a}_t &= (1-\beta) \odot V_{\mathrm{post},t}.
\end{aligned}
\end{equation}
where $I_t$ is the input current, $\beta \in (0,1)^D$ is the learned decay vector, $\vth \in \reals_{>0}^D$ is the fixed threshold vector, $s_t$ is the binary spike, $V_{\mathrm{post},t}$ is the post-reset membrane potential, and $\mathbf{a}_t$ is the leak activation exposed to the next module.

The selective recurrent block itself uses input-dependent membrane parameters and continuous post-reset readout:
\begin{equation}
\label{eq:app_snnblock}
\begin{aligned}
\mathbf{I}_t &= \mathbf{W}_{\mathrm{in}} \mathbf{x}_t, \qquad
\beta_t = \sigmoid\!\big(\mathbf{W}_\beta \mathbf{x}_t + \mathbf{b}_\beta\big), \\
\alpha_t &= \softplus\!\big(\mathbf{W}_\alpha \mathbf{x}_t + \mathbf{b}_\alpha\big), \qquad
\vth(t) = V_{\min} + \big|\mathbf{W}_{\mathrm{th}} \mathbf{x}_t + \mathbf{b}_{\mathrm{th}}\big|, \\
V^{\mathrm{hid}}_{\mathrm{post},t} &= \mathrm{SelectivePLIF}\!\big(\mathbf{I}_t, \beta_t, \alpha_t, \vth(t)\big), \\
\mathbf{o}_t &= \mathbf{W}_{\mathrm{out}} V^{\mathrm{hid}}_{\mathrm{post},t} \odot \mathbf{g}_t + \mathbf{I}_{\mathrm{sk}}.
\end{aligned}
\end{equation}
where $\mathbf{x}_t$ denotes the post-convolution feature vector, $\mathbf{I}_t$ is the projected input current, $\beta_t$, $\alpha_t$, and $\vth(t)$ are the selective decay, input-gain, and threshold vectors, $V^{\mathrm{hid}}_{\mathrm{post},t}$ is the hidden-bank post-reset membrane state, $\mathbf{g}_t$ is the learned multiplicative gate, and $\mathbf{I}_{\mathrm{sk}}$ is the skip path added after gated readout.

The spiking FFN uses two fixed-PLIF branches whose leak activations play the role of multiplicative gate and up channels:
\begin{equation}
\label{eq:app_snnffn}
\begin{aligned}
\mathbf{a}^{(g)}_t &= \phi_{\mathrm{PLIF}}\!\big(\mathbf{W}_g \mathbf{x}_t\big), \\
\mathbf{a}^{(u)}_t &= \phi_{\mathrm{PLIF}}\!\big(\mathbf{W}_u \mathbf{x}_t\big), \\
\mathbf{f}_t &= \mathbf{W}_{\mathrm{down}}\!\big(\mathbf{a}^{(g)}_t \odot \mathbf{a}^{(u)}_t\big) + \mathbf{I}^{\mathrm{ffn}}_{\mathrm{sk}}.
\end{aligned}
\end{equation}
where $\mathbf{W}_g$ and $\mathbf{W}_u$ are the gate and up projections, $\mathbf{a}^{(g)}_t$ and $\mathbf{a}^{(u)}_t$ are their leak activations, $\mathbf{W}_{\mathrm{down}}$ maps the gated product back to the model dimension, and $\mathbf{I}^{\mathrm{ffn}}_{\mathrm{sk}}$ is the FFN skip path.

\subsection{Adaptive temporal aggregation}

Training uses differentiable temporal aggregation weights over the $K$ temporal frames of each token:
\begin{equation}
\label{eq:app_ponder}
\begin{aligned}
p_{s,k} &= \sigmoid\!\big(\mathbf{w}_{\mathrm{halt}}^{\top} \mathbf{z}_{s,k} + b_{\mathrm{halt}}\big), \\
\lambda_{s,k} &= p_{s,k} \prod_{j<k} (1 - p_{s,j}), \\
\bar{\mathbf{z}}_{s} &= \frac{\sum_{k=1}^{K} \lambda_{s,k} \, \mathbf{z}_{s,k}}{\sum_{k=1}^{K}\lambda_{s,k}+\varepsilon}, \qquad
\mathcal{L}_{\mathrm{ponder}} = \sum_{s=1}^{S} \sum_{k=1}^{K} k \, \lambda_{s,k}.
\end{aligned}
\end{equation}
where $\mathbf{z}_{s,k}$ is the framewise sublayer output for token $s$, $p_{s,k}$ is the PonderNet probability at frame $k$, $\lambda_{s,k}$ is the resulting geometric-distribution weight, the denominator normalizes finite-$K$ mass, and $\mathcal{L}_{\mathrm{ponder}}$ regularizes later temporal mass.
The learned temporal-weight distribution is treated as part of the implementation rather than as a separate algorithmic contribution.

\subsection{Output boundary}

After the final decoder layer, the output boundary converts framewise decoder states into token-level logits:
\begin{equation}
\label{eq:app_output}
\begin{aligned}
\mathbf{z}^{\mathrm{out}}_{s,k} &= \phi_{\mathrm{PLIF}}\!\big(\mathrm{RMSNorm}(\mathbf{h}^{(L)}_{s,k})\big), \\
\bar{\mathbf{z}}^{\mathrm{out}}_{s} &= \frac{1}{K} \sum_{k=1}^{K} \mathbf{z}^{\mathrm{out}}_{s,k}, \\
\mathbf{u}_{s} &= \mathbf{W}_{\mathrm{dec}} \bar{\mathbf{z}}^{\mathrm{out}}_{s}, \qquad
\boldsymbol{\ell}_{s} = \mathbf{W}_{\mathrm{emb}} \, \mathrm{DI}(\mathbf{u}_{s}).
\end{aligned}
\end{equation}
where $\mathbf{h}^{(L)}_{s,k}$ is the final-layer framewise residual state, $\bar{\mathbf{z}}^{\mathrm{out}}_{s}$ is the token-level mean after the output PLIF boundary, $\mathbf{W}_{\mathrm{dec}}$ is the decode projection, $\mathrm{DI}(\cdot)$ denotes the divisive-normalization / lateral-inhibition operator, and $\boldsymbol{\ell}_{s}$ are the logits computed with the tied embedding matrix $\mathbf{W}_{\mathrm{emb}}$.

The lateral-inhibition operator is
\begin{equation}
\label{eq:divisive_inhibition}
\mathrm{DI}(\mathbf{u}) =
\boldsymbol{\gamma}_{\mathrm{DI}} \odot
\frac{\mathbf{u}}{\sqrt{\frac{1}{D}\sum_{d=1}^{D} u_d^2 + \varepsilon_{\mathrm{DI}}}},
\end{equation}
where $\boldsymbol{\gamma}_{\mathrm{DI}}$ is a learned gain vector and $\varepsilon_{\mathrm{DI}}$ is a numerical stabilizer.
This operator implements the continuous health-control path referred to in the architecture-stability ablations; it is a divisive-normalization path, not an additional loss term.

\subsection{Training details}
\label{app:train_details}

Structured initialization assigns the hidden SelectivePLIF banks a log-spaced spectrum of initial decay rates so that each visible channel begins with multiple effective timescales.
The threshold scale is matched to a reference temporal horizon to avoid immediate saturation or silence at initialization.

Gradient compensation is applied to the modulation parameters in two phases:
\begin{equation}
\label{eq:app_grad_comp}
\begin{aligned}
\mathbf{g}_{\mathbf{b}_\beta} &\leftarrow \mathbf{g}_{\mathbf{b}_\beta} \oslash \big(\beta \odot (1-\beta)\big), \qquad
\mathbf{g}_{\mathbf{b}_\alpha} \leftarrow \mathbf{g}_{\mathbf{b}_\alpha} \oslash \sigmoid(\mathbf{b}_\alpha), \\
\mathbf{g}^{(\ell)}_{\theta} &\leftarrow
\operatorname{clip}_{[1/c,\,c]}\!\left(
\frac{\big(\prod_{\ell'=1}^{L} \|\mathbf{g}^{(\ell')}_{\theta}\|_2\big)^{1/L}}
{\|\mathbf{g}^{(\ell)}_{\theta}\|_2 + \varepsilon}
\right)\mathbf{g}^{(\ell)}_{\theta},
\end{aligned}
\end{equation}
where $\oslash$ denotes elementwise division, the first line compensates for the local derivatives of the sigmoid and softplus parameterizations, $\theta \in \{\mathbf{b}_\beta, \mathbf{b}_\alpha, \mathbf{b}_{\mathrm{th}}\}$ indexes the modulation parameters participating in the second phase, $\mathbf{g}^{(\ell)}_{\theta}$ is the gradient for parameter group $\theta$ at layer $\ell$, and the clipped geometric-mean rescaling balances gradient norms across layers with compensation limit $c$ and numerical stabilizer $\varepsilon$.

Our fused Triton kernels still execute the full PLIF recurrence sequentially across the $K$ temporal steps; this CUDA path is used for training and capability evaluation, whereas the hardware claims use the separately verified VHDL and FPGA paths reported in Section~\ref{sec:hardware_realization}.

\section{Novelty decomposition and relationship to prior work}
\label{app:novelty}

Table~\ref{tab:novelty} decomposes \textsc{NeuronSpark} into its constituent operations and locates each in the literature, distinguishing genuinely new combinations from known ingredients.
The goal is to be explicit about what is new (the \emph{coupling} and the training-dynamics diagnosis) versus what is adapted from existing work.

\begin{table}[htbp]
\caption{\textbf{Novelty decomposition.} Each component is characterized as: New in this combination, Adapted (known ingredient with a specific modification), or Known (used as-is from prior work).}
\label{tab:novelty}
\centering
\small
\resizebox{\linewidth}{!}{%
\begin{tabular}{llll}
\toprule
Component & Status & Prior art & What is new \\
\midrule
Input-dependent decay $\beta_t$ & Adapted & Mamba~\citep{gu2023mamba} & Applied to PLIF membrane, not continuous SSM \\
Input-dependent threshold $\vth(t)$ & \textbf{New} & N/A & Per-token threshold; resolves firing-regime rigidity \\
Input-dependent gain $\alpha_t$ & \textbf{New} & N/A & Per-token input gain; separate from decay modulation \\
Soft reset & Known & PLIF~\citep{fang2021incorporating} & Used as-is \\
$K$-frame temporal expansion & Adapted & PSN~\citep{fang2024parallel} & PonderNet soft aggregation instead of fixed pooling \\
PonderNet soft aggregation & Adapted & PonderNet~\citep{banino2021pondernet} & Applied to SNN temporal frames, not layer depth \\
PLIF-gated linear-attention memory & Adapted & Fast-weight~\citep{schlag2021linear} & PLIF membrane as scalar write coefficient \\
Gated FFN, Pre-LN residual & Known & LLaMA~\citep{touvron2023llama} & Used as-is \\
RoPE positional encoding & Known & RoFormer~\citep{su2024roformer} & Used as-is \\
RMSNorm & Known & ~\citet{zhang2019root} & Used as-is \\
\midrule
\textbf{Novel combination} & \textbf{New} & N/A & SelectivePLIF + gated linear-attention memory, trained \\
 & & & from random init at 1.1B scale with no backbone transfer \\
\textbf{Firing-regime rigidity diagnosis} & \textbf{New} & N/A & Stable-attractor characterization; globally learnable \\
 & & & $\vth$ insufficient for language modeling \\
\bottomrule
\end{tabular}
}
\end{table}

The central novelty claim is therefore not that any individual operation is new, but that (a) the specific coupling of per-token $\vth(t)$, $\beta_t$, and $\alpha_t$ directly targets the observed attractor problem, (b) the training-dynamics diagnosis (firing-regime rigidity as a dense/silent-regime problem) is new, and (c) the end-to-end training of this combination from random initialization at 1.1B scale is a new demonstration.

\section{Complete training recipe}
\label{app:training_recipe}

\begin{table}[htbp]
\caption{\textbf{Complete training configuration.}}
\label{tab:training_recipe}
\centering
\small
\begin{tabular}{ll}
\toprule
Setting & Value \\
\midrule
Model parameters & 1.16B \\
Layers ($L$) & 24 \\
Model dimension ($D$) & 1024 \\
FFN dimension ($D_{\mathrm{ff}}$) & 3072 \\
Temporal frames ($K$) & 12 \\
State expansion ($N$) & 8 \\
Memory layer period ($m$) & 4 \\
Vocabulary size & 64{,}002 \\
\midrule
Tokenizer & BPE, trained on pretraining corpus \\
Training tokens & 2.1B (reported checkpoint) \\
Context length & 2048 tokens \\
Batch size & 512 sequences (1M tokens/step) \\
Optimizer & AdamW, $\beta_1{=}0.9$, $\beta_2{=}0.95$, $\epsilon{=}10^{-8}$ \\
Learning rate & $3{\times}10^{-4}$, cosine decay to $3{\times}10^{-5}$ \\
Warmup steps & 1000 \\
Weight decay & 0.1 \\
Gradient clipping & 1.0 (global norm) \\
Dropout & 0.0 \\
\midrule
Hardware & 8$\times$ H100 \\
Training time & $\approx$14 days \\
Mixed precision & BF16 \\
Gradient checkpointing & Enabled \\
\midrule
Loss terms & Next-token prediction cross-entropy \\
 & + PonderNet ponder penalty ($\lambda_p = 0.01$) \\
Health-control path & Divisive-normalization / lateral-inhibition operator in \eqref{eq:divisive_inhibition} \\
Checkpoint selection & Final checkpoint (not tuned to validation) \\
Random seed & 42 \\
\midrule
Evaluation & lm-evaluation-harness, zero-shot \\
 & Same protocol as reported baselines \\
\bottomrule
\end{tabular}
\end{table}

\noindent\textbf{Data.} The pretraining corpus is a mixture of publicly available text: approximately 60\% web (filtered Common Crawl subset), 25\% books (Project Gutenberg, Books3 subset), 10\% code (GitHub permissive licenses), and 5\% Wikipedia. No proprietary data is used. Filtering includes language identification, length and duplication filters, license screening for code, and removal of exact benchmark files when identifiable; the anonymized artifact package at \url{https://anonymous.4open.science/r/NeuronSpark} includes dataset-construction scripts, corpus manifests, and reproduction instructions rather than redistributing proprietary or non-public raw data.

\noindent\textbf{Reproducibility scope.} The reported results use the completed large-scale pretraining run with seed 42. The listed configuration is the reported training recipe, and ablation variants diagnose stability under the same architecture-search recipe rather than exhaust all possible retuned variants.

\section{Data-controlled comparison scope}
\label{app:comparison_scope}

Comparing billion-parameter language-model backbones across Transformer, Mamba, and spiking-neural architectures is inherently difficult because the object being compared is a full pretraining run, not a lightweight downstream adapter or a small-scale controlled module.
Achieving absolute data-controlled fairness would require retraining every architecture family from random initialization with the same tokenizer, corpus mixture, data ordering, context length, optimizer schedule, implementation maturity, and compute budget.
This is not a matter of re-running an evaluation harness; each additional control point is another billion-parameter pretraining run, and matching several architecture families across token budgets multiplies that cost by the number of architectures and checkpoints.
Using our reported run as a concrete accounting unit, one \textsc{NeuronSpark}-scale pretraining run costs approximately $8 \times 24 \times 14 \approx 2.7$K H100 GPU-hours; even a minimal three-family controlled comparison at one matched budget would therefore require several additional thousand H100 GPU-hours before considering seeds, intermediate checkpoints, tokenizer variants, or retuning.
Our H100 allocation was rented for the reported pretraining run, while the local 8$\times$RTX 4090 environment was used for evaluation and profiling diagnostics rather than repeated billion-parameter pretraining.
This practical data-control issue is common in large architecture studies: Mamba combines newly trained selective-SSM models with documented comparisons to public open-source language models such as Pythia and RWKV, yet still provides a community-useful architecture reference~\citep{gu2023mamba,biderman2023pythia,peng2023rwkv}.
Our evaluation therefore controls the strongest auditable axes available from public artifacts--parameter scale, released or estimated training-token budget, checkpoint provenance, zero-shot evaluation harness, metric definitions, and task set--but should not be read as a fully data-controlled architecture comparison.
We separate comparable-budget checkpoint comparisons from fully trained capability anchors, with the exact protocol choices reported in Section~\ref{app:eval_columns}.
The hardware comparison uses a separate, vocabulary-aligned derivative of the same 24-layer backbone so that Transformer and Mamba implementations can be matched within 0.323\% and 0.011\% of its 1.260B parameter count.
VHDL results are bit-exact cycle and operation-energy-proxy accounting at separate energy- and speed-oriented points, not routed power or measured joules.
The FPGA result is a physical one-direction timing measurement on one development board, with NeuronSpark using a hybrid Arm/PL path and the baselines using the board's Cortex-A72, so it establishes implementation-level timing rather than compute-resource identity.

\section{Proof of Proposition~\ref{prop:hierarchy}}
\label{app:prop_proof}

We prove both parts of Proposition~\ref{prop:hierarchy} by direct parameter substitution into the SelectivePLIF dynamics \eqref{eq:sel_dyn}.

\noindent\textbf{Part (i): Standard PLIF as a special case.}
For the fixed-PLIF special case, set the modulation projection matrices $\mathbf{W}_\beta = 0$, $\mathbf{W}_\alpha = 0$, $\mathbf{W}_{\mathrm{th}} = 0$ and the biases $\mathbf{b}_\beta = \mathrm{logit}(\beta)$, $\mathbf{b}_\alpha = \log(\exp(\alpha_0)-1)$ (softplus inverse), $\mathbf{b}_{\mathrm{th}} = \vth - V_{\min}$ for target static values $\beta \in (0,1)$, $\alpha_0 = 1-\beta$, $\vth > V_{\min}$.
Then for any input $\mathbf{h}_t$:
\[
  \beta_t = \sigmoid(\mathbf{b}_\beta) = \beta, \quad
  \alpha_t = \softplus(\mathbf{b}_\alpha) = 1-\beta, \quad
  \vth(t) = V_{\min} + |\mathbf{b}_{\mathrm{th}}| = \vth.
\]
Substituting into \eqref{eq:sel_dyn}: $V_t = \beta \cdot V_{t-1} + (1-\beta) \cdot I_t$, $s_t = \Theta(V_t - \vth)$, $V_t \leftarrow V_t - \vth \cdot s_t$, which is exactly \eqref{eq:plif_base}. \hfill$\square$

\noindent\textbf{Part (ii): LTI SSM as a further special case.}
Starting from Part (i), take the no-spike limit $\vth \to \infty$.
Then $\Theta(V_t - \vth) = 0$ for all finite $V_t$, so $s_t = 0$ identically and the soft-reset term $\vth \cdot s_t = 0$ vanishes.
The dynamics reduce to $V_t = \beta \cdot V_{t-1} + (1-\beta) \cdot I_t$, which is a scalar first-order linear recurrence with real-valued decay $\beta \in (0,1)$, the defining equation of a scalar LTI SSM with decay $\beta$ and input gain $(1-\beta)$. \hfill$\square$

\noindent\textbf{Strictness of inclusions.}
At the membrane-transition level, standard PLIF with fixed $\beta, \vth$ cannot implement input-dependent forgetting: for any two distinct inputs $\mathbf{h}_t \neq \mathbf{h}_{t'}$ the coefficient is identical ($\beta_t = \beta_{t'} = \beta$).
SelectivePLIF with non-constant $\mathbf{W}_\beta$ produces distinct transitions.
Hence standard PLIF $\subsetneq$ SelectivePLIF.
The LTI SSM $\subsetneq$ standard PLIF inclusion is strict since standard PLIF includes a spiking nonlinearity absent from the LTI SSM.

\section{Detailed mathematical derivations}
\label{app:math}

\subsection{Affine decomposition and fused execution for PLIF dynamics}

The PLIF dynamics \eqref{eq:sel_dyn} can be decomposed into a linear recurrence coupled with non-linear spike detection.
The linear component $V_t = \beta_t \cdot V_{t-1} + u_t$ is an affine map $f_t(x) = \beta_t \cdot x + u_t$, and the composition of $K$ such maps can be computed in $O(K \log K)$ work and $O(\log K)$ depth via the Hillis-Steele parallel prefix scan.
The composition rule for two affine maps is given in \eqref{eq:compose}:
\begin{equation}
\label{eq:compose}
  (a_2, b_2) \circ (a_1, b_1) = (a_2 \cdot a_1, \; a_2 \cdot b_1 + b_2),
\end{equation}
where $(a_1, b_1)$ and $(a_2, b_2)$ represent affine maps $f_1(x) = a_1 x + b_1$ and $f_2(x) = a_2 x + b_2$ respectively, and $(a_2 \cdot a_1, \; a_2 \cdot b_1 + b_2)$ is their composition $f_2 \circ f_1$.

This affine-scan view is useful for understanding the linear subproblem, but it does not directly solve the full PLIF recurrence because spike detection and soft reset introduce stepwise data dependencies.
For the full PLIF dynamics including spike and soft reset, our fused Triton kernel therefore performs a sequential scan across $K$ steps, while parallelizing across all $\mathrm{batch} \times \mathrm{width}$ columns simultaneously.
The backward pass uses reverse accumulation with reconstructed $V_{\mathrm{pre},t} = V_{\mathrm{post},t} + \vth(t) \cdot s_t$ and computes the sigmoid surrogate gradient inline.

\subsection{Surrogate gradient for PLIF}

The backward pass through the spike function $s_t = \Theta(V_t - \vth)$ uses a sigmoid surrogate~\citep{neftci2019surrogate}, as shown in \eqref{eq:surrogate}:
\begin{equation}
\label{eq:surrogate}
  \frac{\partial s_t}{\partial V_t} \approx \alpha_{\mathrm{sg}} \cdot \sigma(\alpha_{\mathrm{sg}} \cdot (V_t - \vth)) \cdot (1 - \sigma(\alpha_{\mathrm{sg}} \cdot (V_t - \vth))),
\end{equation}
where $\alpha_{\mathrm{sg}}$ is the surrogate gradient sharpness parameter (we use $\alpha_{\mathrm{sg}} = 4.0$), and $\sigma(\cdot)$ denotes the sigmoid function.
This approximation provides smooth, non-zero gradients in a neighborhood of the threshold, enabling effective backpropagation through the spiking nonlinearity.

\section{Additional experimental details}
\label{app:details}

\subsection{Model configurations, optimization, and data}
\label{app:exp_setup}

Table~\ref{tab:configs} summarizes the checkpoint evaluated in the main text.
$D$ denotes the model dimension, $D_{\mathrm{ff}}$ the FFN intermediate dimension, $L$ the number of layers, $K$ the maximum number of SNN temporal frames, and $N$ the state expansion factor.

\begin{table}[ht]
  \caption{\textbf{Model configurations.}}
  \label{tab:configs}
  \centering
  \small
  \begin{tabular}{lcccccccc}
    \toprule
    Model & $D$ & $D_{\mathrm{ff}}$ & $L$ & Mem.\ int. & $K$ & $N$ & Params & Tokens \\
    \midrule
    \textsc{NeuronSpark}-1.1B & 1024 & 3072 & 24 & 4 & 12 & 8 & 1.16B & 2.1B \\
    \bottomrule
  \end{tabular}
\end{table}

The evaluated model has 18 SNN decoder layers and 6 associative-memory layers, with memory layers inserted every fourth layer.
The vocabulary contains 64{,}002 tokens.
The reported checkpoint consumes 2.1B training tokens.
The large-scale pretraining run used a rented 8$\times$H100 allocation and stopped when that rental window expired; VHDL validation and the physical FPGA timing run were conducted separately from this training allocation.
The available large-scale TensorBoard trace reaches a terminal smoothed training loss of approximately 3.5 and is used for the trainability diagnostic in Figure~\ref{fig:training_diagnostics}.

\paragraph{Training and evaluation summary.}
Table~\ref{tab:hyperparams} lists the run settings that determine the reported results.

\begin{table}[ht]
  \caption{\textbf{Training and evaluation summary.}}
  \label{tab:hyperparams}
  \centering
  \small
  \begin{tabular}{ll}
    \toprule
    Item & Value \\
    \midrule
    Training objective & Next-token prediction \\
    Reported checkpoint & \textsc{NeuronSpark}-1.1B \\
    Training tokens & 2.1B \\
    Available large-scale trace endpoint & 479K steps \\
    Terminal smoothed loss in available trace & 3.5 \\
    Precision & bf16 \\
    Surrogate gradient $\alpha_{\mathrm{sg}}$ & 4.0 \\
    $V_{\min}$ (threshold floor) & 0.1 \\
    Evaluation & lm-evaluation-harness zero-shot \\
    Hardware checkpoint & Vocabulary-aligned 24-layer derivative, 1.260B parameters \\
    VHDL baselines & Transformer 1.264B; Mamba 1.260B \\
    FPGA platform & ALINX VD100 / AMD Versal AI Edge VE2302 \\
    FPGA precision and prompt & BF16; 2 prompt tokens; 4 generated tokens \\
    \bottomrule
  \end{tabular}
\end{table}

\subsection{Training traces and stability diagnostics}
\label{app:training_diagnostics}

The Hugging Face TensorBoard scalar trace covers steps 31--479{,}149, showing loss decreasing from 11.24 into the low-loss training band.
The last single-batch scalar is noisy, while the terminal moving average remains near 3.5.
Development stability curves use short 4090 FSDP trials rather than full pretraining runs.
They were stopped once a variant stagnated in the 7--8 loss band, reversed after an initial decrease, or produced NaNs, so they serve as rejection diagnostics for unstable design choices rather than scaled training comparisons.
The accepted architecture column reports the matched-step accepted-reference loss used in Figure~\ref{fig:training_diagnostics}(b); the full 1.16B H100 pretraining trace is shown separately in Figure~\ref{fig:training_diagnostics}(a).
The short-run conclusion is negative and operational: none of the rejected variants produced a trajectory that continued from early loss reduction into the final low-loss regime reached by the accepted architecture.
Here the health-control path denotes the continuous lateral-inhibition / divisive-normalization operator defined in \eqref{eq:divisive_inhibition}.
The Sinkhorn-health variant replaces this continuous normalization signal by a Sinkhorn-projected health score, while the HC-$\alpha$ variant decouples the input-gain parameterization from this control path.

\begin{table}[ht]
  \caption{\textbf{Matched-step development-stability runs.}}
  \label{tab:stability_runs}
  \centering
  \scriptsize
  \setlength{\tabcolsep}{3.0pt}
  \resizebox{\linewidth}{!}{%
  \begin{tabular}{lccccl}
    \toprule
    Variant & End step & Final loss & Best loss & Accepted ref. & Rejection signal \\
    \midrule
    MPD-AGL w/o phase-2 & 4.79K & 7.24 & 6.78 @ 3.76K & 6.86 & Reversed after early descent \\
    w/o input-dependent PLIF & 5.08K & 7.69 & 7.06 @ 3.83K & 6.85 & Reversed after early descent \\
    HC $\alpha$ decoupled & 0.16K & 7.72 & 7.60 @ 0.15K & 7.68 & No sustained descent \\
    Sinkhorn health & 2.05K & 7.60 & 7.41 @ 1.19K & 7.14 & Stayed in high-loss band \\
    Cortical lateral & 4.11K & 7.68 & 7.25 @ 1.19K & 6.89 & Reversed after early descent \\
    \bottomrule
  \end{tabular}
  }
\end{table}

\begin{table}[ht]
  \caption{\textbf{Rejected-variant definitions.}}
  \label{tab:ablation_definitions}
  \centering
  \scriptsize
  \setlength{\tabcolsep}{4.0pt}
  \begin{tabular}{p{0.18\linewidth}p{0.38\linewidth}p{0.36\linewidth}}
    \toprule
    Variant & What changed & Diagnostic purpose \\
    \midrule
    MPD-AGL w/o phase-2 & Removed cross-layer equalization from modulation-gradient compensation & Tests whether local compensation alone stabilizes selective PLIF parameters \\
    w/o input-dependent PLIF & Replaced token-dependent PLIF modulation by fixed-parameter PLIF dynamics & Tests whether neuron-level input dependence is required for stable membrane updates \\
    HC $\alpha$ decoupled & Used a separate $\alpha$ parameterization from the health-control path & Tests whether input gain should be decoupled from health-control modulation \\
    Sinkhorn health & Replaced the health signal by a Sinkhorn-projected health score & Tests whether projected global health balancing stabilizes neuron utilization \\
    Cortical lateral & Added cross-token causal spike propagation & Tests whether cortical-style lateral propagation helps contextual exchange \\
    \bottomrule
  \end{tabular}
\end{table}

\FloatBarrier
\subsection{VHDL and FPGA hardware protocols}
\label{app:system_profile}

Table~\ref{tab:vhdl_profile} reports the raw context-2048 values behind Figure~\ref{fig:vhdl_costs}.
The Transformer and Mamba widths are selected to match the 1,260,062,732-parameter hardware checkpoint within 1\%; NeuronSpark's energy and speed rows are separately accepted operating points derived from integrated event execution, hierarchical zero/tile gating, and folded-product page caching.

\begin{table}[ht]
  \caption{\textbf{Same-parameter VHDL cost accounting.}}
  \label{tab:vhdl_profile}
  \centering
  \scriptsize
  \setlength{\tabcolsep}{4.0pt}
  \begin{tabular}{lrrr}
    \toprule
    Datapath & Parameters & Energy proxy & Cycles \\
    \midrule
    Transformer, matched & 1,264,129,912 & $2.768430{\times}10^{12}$ & 42,242,890 \\
    Mamba, matched & 1,260,206,464 & $0.829721{\times}10^{12}$ & 12,660,736 \\
    \textsc{NeuronSpark}, energy point & 1,260,062,732 & $0.273821{\times}10^{12}$ & 17,463,296 \\
    \textsc{NeuronSpark}, speed point & 1,260,062,732 & $0.273847{\times}10^{12}$ & 2,496,512 \\
    \bottomrule
  \end{tabular}
\vspace{0.15em}
\begin{flushleft}
\scriptsize Energy is a circuit-aware operation proxy normalized to dense BF16 operations; it is not routed FPGA power. All rows use a 2,048-token context and pass their declared VHDL checks.
\end{flushleft}
\end{table}

Table~\ref{tab:fpga_profile} reports the physical-board values behind Figure~\ref{fig:fpga_speed}.
All rows use BF16, a two-token prompt, four generated tokens, tracing disabled, and the same VD100 development board carrying an AMD Versal AI Edge VE2302 device.

\begin{table}[ht]
  \caption{\textbf{Physical FPGA inference measurements.}}
  \label{tab:fpga_profile}
  \centering
  \scriptsize
  \setlength{\tabcolsep}{5.0pt}
  \begin{tabular}{llrr}
    \toprule
    Model & Execution path & TTFT (s) & Decode TPS \\
    \midrule
    Transformer & Cortex-A72 & 73.509 & 0.024262 \\
    Mamba & Cortex-A72 & 70.683 & 0.024125 \\
    \textsc{NeuronSpark} & Hybrid Arm/PL & 24.000 & 0.082440 \\
    \bottomrule
  \end{tabular}
\vspace{0.15em}
\begin{flushleft}
\scriptsize All four measured baseline and repeated-run records have \texttt{measurement\_valid=1}; the maximum A/B spread is 0.337\%. This is a same-board implementation comparison, not an identical-compute-resource comparison.
\end{flushleft}
\end{table}

\FloatBarrier
\subsection{Neural dynamics diagnostic details}
\label{app:dynamics_details}

\begin{figure}[ht]
  \centering
  \includegraphics[width=0.58\linewidth]{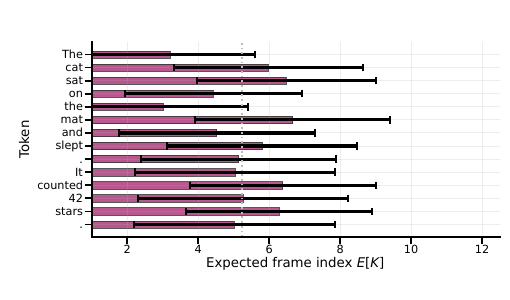}
  \caption{\textbf{Per-token soft temporal-depth diagnostic.}}
  \label{fig:token_depth_appendix}
\end{figure}

The token-depth diagnostic in Figure~\ref{fig:neural_dynamics_v2} uses 64 TinyStories validation prompts and 4096 token items.
Buckets are surface/lexical heuristics over tokenizer outputs rather than externally supervised part-of-speech labels.
Figure~\ref{fig:token_depth_appendix} shows the per-token bar diagnostic behind the compact heatmap.
Values are expected frame indices over the executed $K{=}12$ frames, not fractional executed steps.
Table~\ref{tab:depth_bucket_stats} reports the corresponding distributional summaries.

\begin{table}[ht]
  \caption{\textbf{Token-category soft temporal-depth statistics.}}
  \label{tab:depth_bucket_stats}
  \centering
  \scriptsize
  \setlength{\tabcolsep}{3.0pt}
  \begin{tabular}{lrrrrrr}
    \toprule
    Bucket & $n$ & Mean & Std & Median & P10 & P90 \\
    \midrule
    Articles/determiners & 507 & 3.84 & 0.70 & 3.65 & 3.14 & 4.87 \\
    Pronouns & 221 & 5.01 & 0.28 & 5.05 & 4.60 & 5.28 \\
    Prep./conj./particles & 468 & 4.41 & 0.71 & 4.31 & 3.61 & 5.45 \\
    Aux./copula/modal & 242 & 4.87 & 0.47 & 4.76 & 4.32 & 5.62 \\
    Content word starts & 1982 & 5.80 & 0.58 & 5.87 & 4.94 & 6.47 \\
    Subword pieces & 114 & 5.73 & 0.40 & 5.77 & 5.25 & 6.21 \\
    Punctuation/quotes & 562 & 4.71 & 0.86 & 5.00 & 3.50 & 5.60 \\
    \bottomrule
  \end{tabular}
\end{table}

\FloatBarrier
\subsection{Evaluation columns}
\label{app:eval_columns}

The main benchmark comparison reports ARC-E, ARC-C, WinoGrande, BoolQ, PIQA, OpenBookQA, MMLU, and C-Eval.
We report \texttt{acc\_norm} where the harness provides it for ARC-E/C, PIQA, and OpenBookQA, and \texttt{acc} for the remaining tasks.
The Mamba-1.6B rows use the public LightOn Mambaoutai checkpoints from \texttt{lightonai/mambaoutai}.
The pre-1000 row is the closest released low-budget Mamba checkpoint, with its token count estimated as $1000 \times 896 \times 4096 \approx 3.67$B tokens from the released training configuration.

For the SNN-specific comparison in Table~\ref{tab:benchmark_results}, SpikeGPT values use our evaluation of the official 216M OpenWebText checkpoint where available.
For CSSA/SKD SpikingLLM, no public checkpoint is available for re-evaluation, and the paper reports both OPT-1.3B reference scores and SNN/OPT relative performance.
We therefore compute each displayed CSSA/SKD value as
\begin{equation}
\label{eq:cssa_score_normalization}
\begin{aligned}
s_{\mathrm{CSSA},j}^{\mathrm{shown}}
&=
s_{\mathrm{OPT},j}^{\mathrm{ours}}
\cdot
\frac{s_{\mathrm{CSSA},j}^{\mathrm{paper}}}
     {s_{\mathrm{OPT},j}^{\mathrm{paper}}},
\end{aligned}
\end{equation}
where $j$ indexes the downstream task, $s_{\mathrm{CSSA},j}^{\mathrm{paper}}$ and $s_{\mathrm{OPT},j}^{\mathrm{paper}}$ are the CSSA/SKD SpikingLLM and OPT-1.3B scores reported in the SpikingLLM paper, and $s_{\mathrm{OPT},j}^{\mathrm{ours}}$ is our lm-evaluation-harness reproduction of OPT-1.3B under the protocol used for Table~\ref{tab:benchmark_results}.
This keeps the reported SNN/OPT ratio from the SpikingLLM paper while placing the absolute score on our OPT-1.3B evaluation anchor.
WTA-SpikingFormer values are taken from the reported WD causal-decoder commonsense setting, except SST-2, which is the reported WE encoder-only value.

%%%%%%%%%%%%%%%%%%%%%%%%%%%%%%%%%%%%%%%%%%%%%%%%%%%%%%%%%%%%
\FloatBarrier

\end{document}